\documentclass[prx,nofootinbib,onecolumn,showkeys,groupaddress,preprintnumbers,floatfix,
superscriptaddress]{revtex4-1}

\usepackage{etex}                        \usepackage{ifpdf}                       \usepackage{xspace}                      \usepackage[table]{xcolor}               \usepackage{setspace}                    

\PassOptionsToPackage{final}{graphicx}   \usepackage{graphicx}                    

\makeatletter                            \@ifclassloaded{beamer}{}{\PassOptionsToPackage{pagebackref}{hyperref}}
\makeatother
\usepackage{hyperref}
\usepackage{url}                         \usepackage[super]{nth}                  \usepackage{contour}                     \usepackage{verbatim}                    \usepackage{ragged2e}                    \usepackage{attrib}                      \usepackage{adjustbox}                   \usepackage[absolute,overlay]{textpos}   \usepackage{calc}                        

\usepackage{amsmath}                     \usepackage{amscd}                       \usepackage{amsfonts}                    \usepackage{amssymb}                     \usepackage{amsthm}                      \usepackage{scalerel}                    \usepackage{bm}                          \usepackage{bbm}                         \usepackage{braket}				               \usepackage{mathtools}                   \usepackage{etoolbox}                    \usepackage{textcomp}                    \usepackage{stmaryrd}                    \usepackage{nicefrac}                    \usepackage[binary-units]{siunitx}       

\usepackage[T1]{fontenc}                 \usepackage{lmodern}                     \usepackage[scaled=0.72]{beramono}       \usepackage{microtype}                   

\usepackage{algorithm}                   \usepackage{algorithmicx}                

\usepackage{booktabs}                    \usepackage{dcolumn}                     

\makeatletter
  \@ifclassloaded{revtex4-1}
  {
    \usepackage[caption=false]{subfig}   }
  {
    \usepackage{caption}
    \usepackage{subcaption}
  }
  \@ifclassloaded{beamer}
  {
    \usepackage[backend=biber,
                autocite=superscript,
                doi=false,
                url=false,
                sorting=none]{biblatex}
    \usepackage[useregional,showdow]{datetime2}
  }
  {
\usepackage{enumitem}                }
\makeatother

\definecolor{ryan}{RGB}{64, 0, 64}
\definecolor{nix}{RGB}{255, 0, 0}

\definecolor{ucdblue1}{cmyk}{.87,.46,0,.49} \definecolor{ucdblue2}{cmyk}{1., .56, 0., .34}
\colorlet{ucdblue}{ucdblue2}

\makeatletter
\def\@lox@prtc{\section*{\@fxlistfixmename}\begingroup\def\@dotsep{4.5}}
\def\@lox@psttc{\endgroup}
\makeatother

\usepackage{epigraph}

\colorlet {past_color}    {red}
\colorlet {pres_color}    {blue}
\colorlet {futu_color}    {black!30!green}

\colorlet {temp_color_1}  {red!50!blue}
\colorlet {temp_color_2}  {red!50!green}
\colorlet {temp_color_3}  {blue!50!green}

\colorlet {hmu_color}     {blue!67!green}
\colorlet {rhomu_color}   {temp_color_1!80!blue}
\colorlet {rmu_color}     {blue}
\colorlet {bmu_1_color}   {temp_color_1}
\colorlet {bmu_2_color}   {temp_color_3}
\colorlet {qmu_color}     {temp_color_1!67!green}
\colorlet {wmu_color}     {temp_color_2!57!blue}
\colorlet {sigmamu_color} {temp_color_2}

\usepackage{listings}
\lstdefinestyle{mypython}{
language=Python,                        basicstyle=\small\ttfamily,             keywordstyle=\color{green!50!black},    commentstyle=\color{gray},              numbers=left,                           numberstyle=\tiny,                      stepnumber=1,                           numbersep=5pt,                          backgroundcolor=\color{gray!10},        frame=none,                             tabsize=2,                              captionpos=b,                           breaklines=true,                        breakatwhitespace=false,                showspaces=false,                       showtabs=false,                         morekeywords={as},                      }

\theoremstyle{plain}    
\theoremstyle{plain}    
\theoremstyle{plain}    
\theoremstyle{plain}    
\theoremstyle{plain}    
\theoremstyle{plain}    
\theoremstyle{plain}    
\theoremstyle{plain}    
\theoremstyle{plain}    
\theoremstyle{plain}    
\theoremstyle{plain}    \newtheorem{Def}{Definition}
\theoremstyle{plain}

\newcommand{\CausalState}       { \mathcal{S} }

\newcommand{\forward}{+}
\newcommand{\reverse}{-}
\newcommand{\forwardreverse}{\pm} 

\newcommand{\FutureCausalState} { {\CausalState}^{\forward} }

\newcommand{\PastCausalState}   { {\CausalState}^{\reverse} }

\newcommand{\lastindex}[2]{
  \edef\tempa{0}
  \edef\tempb{#2}
  \ifx\tempa\tempb
\edef\tempc{#1}
  \else
\edef\tempa{0}
    \edef\tempb{#1}
    \ifx\tempa\tempb
      \edef\tempc{#2}
    \else
      \edef\tempc{#1+#2}
    \fi
  \fi
  \tempc
}

\newcommand{\CSjoint}[1][,]{
   \edef\tempa{:}
   \edef\tempb{#1}
   \ifx\tempa\tempb
\ensuremath{\FutureCausalState\!#1\PastCausalState}
   \else
\ensuremath{\FutureCausalState#1\PastCausalState}
   \fi
}
\newcommand{\CSjointKL}[3][,]{
   \edef\tempa{:}
   \edef\tempb{#1}
   \ifx\tempa\tempb
\ensuremath{\FutureCausalState_{#2}\!#1\PastCausalState_{#3}}
   \else
\ensuremath{\FutureCausalState_{#2}#1\PastCausalState_{#3}}
   \fi
}

\newif\ifpm
\edef\tempa{\forwardreverse}
\edef\tempb{\pm}
\ifx\tempa\tempb
   \pmfalse
\else
   \pmtrue
\fi

\ifcsname mathclap\endcsname
  \relax
\else
  \def\clap#1{\hbox to 0pt{\hss#1\hss}}

\fi

\newcommand{\op} [3] [] {
  \ensuremath{
    \operatorname{#2_{#1}}
    \if\relax\detokenize{#3}\relax
    \else
      \left[ #3 \right]
    \fi
  }
  \xspace
}

\renewcommand{\H}  [2] []       {\op[#1]{H}{#2}}

\makeatletter
  \@ifclassloaded{beamer}
  {
    \hypersetup{
      linkcolor={blue!50!black},
      citecolor={blue!50!black},
      urlcolor={blue!80!black},
    }
  }
  {
    \hypersetup{
      colorlinks,
      linkcolor={blue!50!black},
      citecolor={blue!50!black},
      urlcolor={blue!80!black},
    }
  }
\makeatother

\usepackage{empheq}
\usepackage{overpic}
\usepackage{mathrsfs}

\newcommand{\St}{\mathcal{S}}

\newcommand{\SSet}{\boldsymbol{\mathcal{S}}}
\newcommand{\Abet}{\mathcal{X}}

\newcommand{\stationary}{\boldsymbol{\pi}}

\renewcommand{\H}{\operatorname{H}}

\newcommand{\Wstate}[1][c]{#1}

\newcommand{\params}{\vec{\theta}}
\newcommand{\est}{\hat{\params}_L}
\newcommand{\FI}[1][L]{\boldsymbol{\mathcal{F}}_{#1}}
\newcommand{\cov}[1]{\text{var} \Bigl( {#1} \Bigr)}
\newcommand{\dF}[1][L]{\boldsymbol{f}_{#1}}
\newcommand{\FIrate}{\boldsymbol{f}}
\newcommand{\EFI}[1][L]{\boldsymbol{\epsilon}_{#1}}
\newcommand{\MarkovOrder}{\mathscr{M}}

\newcommand{\Blackwell}{\stationary_{\params}}
\newcommand{\ones}{\boldsymbol{1}}

\newcommand{\Tones}{\ones}
\newcommand{\Wones}{\ket{\ones}}

\newcommand{\Tpi}{\stationary_T}
\newcommand{\Wpi}{\bra{\stationary}}

\newcommand{\co}{\mathscr{K}}
\newcommand{\Pro}{\text{P}_{\params}}
\newcommand{\es}{\varepsilon}
\newcommand{\ES}{\mathcal{E}}
\newcommand{\esSet}{\boldsymbol{\mathcal{E}}}
\newcommand{\StartMS}{\bra{\Wstate[\es(\varnothing)]}}

\makeatletter
\@booleanfalse\titlepage@sw
\makeatother

\begin{document}

\def\ourTitle{Ultimate limit on learning
	non-Markovian 
	behavior:\\Fisher 
	information rate and excess information
}

\def\ourAbstract{We address
the fundamental limits of learning 
unknown parameters of any 
stochastic process from time-series data,
and discover exact closed-form expressions for how optimal inference scales with observation length.
Given a parametrized class of candidate models, the Fisher information of observed sequence probabilities 
lower-bounds the variance in model estimation from finite data.
As sequence-length increases,
the minimal variance 
scales as the square inverse of the length---with constant coefficient given by the 
information rate.
We 
discover 
a simple closed-form expression for
this information rate, 
even in the case of
infinite Markov order. 
We furthermore obtain the exact analytic lower bound on model variance
from the 
observation-induced
metadynamic among belief states. 
We 
discover ephemeral, exponential, and more general 
modes of convergence to the asymptotic information rate.
Surprisingly, 
this
myopic information rate
converges to the asymptotic
Fisher
information rate
with 
exactly
the 
same relaxation timescales 
that appear
in
the myopic entropy rate
as it converges
to the 
Shannon entropy rate for the process.
We illustrate these results
with
a sequence of examples 
that highlight
qualitatively distinct features of stochastic processes that 
shape optimal
learning.
}

\def\ourKeywords{}

\hypersetup{
  pdfauthor={Paul M. Riechers},
  pdftitle={\ourTitle},
  pdfsubject={\ourAbstract},
  pdfkeywords={\ourKeywords},
  pdfproducer={},
  pdfcreator={}
}

\title{\ourTitle}

\author{Paul M. Riechers}
\email{pmriechers@gmail.com}

\affiliation{Beyond Institute for Theoretical Science,
San Francisco, California, USA}

\date{\today}
\bibliographystyle{unsrt}

\begin{abstract}
\ourAbstract
\end{abstract}

\date{\today}
\maketitle

\setstretch{1.1}

The longer you know someone, the more easily you can predict what they will say or how they will act in a certain situation.
Predictive artificial intelligence (AI) succeeds similarly
by modeling the correlations in language (for large language models like GPT-4) or modeling  correlations in more general data streams (e.g., for an AI assessing road conditions and proper behavior as it drives).  For such stochastic processes, longer training sequences enable discovery of hidden structure
and appropriate behavior as the inferred model becomes closer to the statistical ground truth.

Yet---despite routine  model inference of correlated data throughout science and industry---surprisingly
little is known about the fundamental limits on this type of learning. 
How well can learning improve with longer training sequences in general?

Building upon the recent framing of this learning problem~\cite{Radaelli22_Fisher},
we discover in closed form how the Fisher information of sequence probabilities scales with increasing sequence length---which fundamentally bounds how well 
the parameters of a model can be inferred 
from data.
Our answers
leverage 
methods developed in
the field of computational mechanics, which has extensively studied 
optimal prediction
of any known source of complex behavior
~\cite{Crut12a, Barn13a, Riec18_SSAC1}.
This connection is surprising because the current question is about learning an \emph{unknown} model, rather than synchronizing to a known model.

The learning problem assumes a parametrized family of stochastic processes.
The parameter vector $\params \in \boldsymbol{\Theta}$ induces a joint probability distribution over observables
$\Pr_{\params}(X_{1:L})$
for any sequence-length $L$,
where 
 $X_{1:L} = X_1, X_2, \dots , X_L$ denotes the sequence of random variables.  Similarly $x_{1:L} = x_1, x_2, \dots , x_L$ will denote a realization of this sequence.
The random variable $X_\ell$ can take on values in the observable alphabet $\mathcal{X}$,
which we assume to be countable.
Our derivation furthermore assumes 
a discrete domain (e.g., discrete time).
However, the continuous-time regime can be obtained in the appropriate limit of vanishingly small time steps.

\section{Introduction to Fisher information rate}

The Fisher information matrix $\FI$ bounds the auto-covariance of any estimator $\est$ of the process, biased or unbiased, estimated from a single length-$L$ sequence.
For an unbiased estimator, the Cramer--Rao bound takes the especially simple form:
\begin{align}
	\cov{\est} \geq \FI^{-1} ~,
\end{align}	
where the matrix inequality means that
$\cov{\est} - \FI^{-1} $ is positive semidefinite.
The auto-covariance matrix
can be written more explicitly as
$\cov{\est} = \Bigl\langle \est \, \est^{\, \top} 
\Bigr\rangle  -  \Bigl\langle \est \Bigr\rangle \Bigl\langle \est^{\, \top} \Bigr\rangle$.
With
a single parameter, 
this
reduces to the simpler scalar statement that the variance of the estimated parameter must be at least as large as the reciprocal of the Fisher information.

The Fisher information matrix in this context has elements
\begin{align}
(\FI)_{m, n}	
= 
\Bigl\langle  \bigl[ \partial_{\theta_m} \ln \Pr_{\params}(X_{1:L})  \bigr]
\bigl[ \partial_{\theta_n} \ln \Pr_{\params}(X_{1:L})   \bigr]
\Bigr\rangle_{\Pr_{\params}(X_{1:L})} ~,
\end{align}	
where $\theta_m$ and $\theta_n$ are elements of the 
parameter vector $\vec{\theta}$.
Under the standard regularity assumptions,
this is equal to the more useful form:
\begin{align}
	(\FI)_{m, n}	= - \bigl\langle  \partial_{\theta_m} \partial_{\theta_n}   \ln \Pr_{\params}(X_{1:L})  
	\bigr\rangle_{\Pr_{\params}(X_{1:L})} ~.
\end{align}	
The Fisher information matrix is symmetric and positive semidefinite.

Since 
$\Pr_{\params}(X_{1:L} = x_{1:L}) = \Pr_{\params}(X_1 = x_1) \Pr_{\params}( X_2 = x_2 | X_1 = x_1) \dots \Pr_{\params}(X_L = x_L | X_{1:L-1} =  x_{1: L-1})$,
the
Fisher information decomposes 
such that 
the difference between Fisher information matrices of subsequent lengths,
\begin{align}
	\dF \coloneqq \FI - \FI[L-1] ~,
	\label{eq:dFdef}
\end{align}	
is itself a valid Fisher information matrix with elements
\begin{align}
(\dF)_{m,n}
= - \bigl\langle  \partial_{\theta_m} \partial_{\theta_n}   \ln \Pr_{\params}(X_L | X_{1:L-1})  
\bigr\rangle_{\Pr_{\params}(X_{1:L})} 
\label{eq:MyopicInfoExpression}
\end{align}	
for $L \geq 2$.
We define $\FI[0] = 0$
so that $\dF[1] = \FI[1]$
with elements $(\FI[1])_{m,n} = - \bigl\langle  \partial_{\theta_m} \partial_{\theta_n}   \ln \Pr_{\params}(X_{1})  
\bigr\rangle_{\Pr_{\params}(X_{1})} $.
We call $\dF$ the `myopic information rate', in analogy with the 
`myopic entropy rate' of the process~\cite{Riec18_SSAC1}.
From Eq.~\eqref{eq:dFdef},
it should be clear that 
the net Fisher information
for length-$L$ sequences is
$\FI = \sum_{\ell=1}^{L} \FIrate_\ell$.

For stationary stochastic processes,
Ref.~\cite{Radaelli22_Fisher} recently
introduced the concept of the 
\emph{information rate} 
$\FIrate \coloneqq \lim_{L \to \infty} \dF$
and the 
\emph{excess information}
$\boldsymbol{\epsilon} \coloneqq  \lim_{L \to \infty} \EFI $,
where we define the \emph{myopic excess 
information} as $\EFI \coloneqq \FI - L \FIrate $.
Ref.~\cite{Radaelli22_Fisher} showed that the 
excess information can be either 
positive or negative.
Ref.~\cite{Radaelli22_Fisher} 
also showed
that, for processes of finite Markov order
$\MarkovOrder$,
the
myopic excess information saturates at the Markov order: 
$\EFI[L] = \EFI[\MarkovOrder] = \boldsymbol{\epsilon}$ for $L \geq \MarkovOrder$.
Recall that
the Markov order can be defined 
for a stationary stochastic process
as
$\MarkovOrder = \min \bigl\{ L \in \mathbb{N}_0 :  \Pr_{\params}(X_0 | X_{-L:-1}) = \Pr_{\params}(X_0 | X_{-\infty:-1})  \bigr\}$.
However, stochastic processes typically have infinite Markov order~\cite{Jame10a}.

The information rate is important because it controls the asymptotic behavior of the variance of any estimator of the process
as longer sequences are taken into account:
\begin{align}
	\FI = \EFI + L \FIrate ~.
\end{align}	
Accordingly, whenever the 
excess information $\boldsymbol{\epsilon}$ is finite,
the Fisher information in a long sequence scales exactly as the information rate
\begin{align}
	\lim_{L \to \infty} \frac{\FI}{L}  = \FIrate ~.
\end{align}

Here we give the first closed-form expressions for the information rate and the excess information.
Our relations are valid
whether the process has finite or infinite Markov order.
Moreover, we will show that 
myopic \emph{information} rates 
converge with exactly the same length dependence
as 
myopic \emph{entropy} rates 
for any stochastic process.
This 
elucidates their relationship with various known entropic quantities, and 
allows us to determine
when $\EFI$ will be positive or negative,
and when $\boldsymbol{\epsilon}$ will be finite.

\section{Preview of our results}

Any stochastic process
has infinitely many 
latent-state representations---each called a `presentation' of the process~\cite{Crut10a}.
If we allow infinite-state models, then every process has a representation as 
a unifilar hidden Markov model (HMM).
One of our major results is that
\emph{the information rate can be calculated 
	directly from any unifilar presentation of 
	the stochastic process}, 
so long as 
it
faithfully represents the parametrized process
in a neighborhood of $\params$.
\begin{Def}
A latent-state model of a process is \emph{unifilar} 
if the current state and next observable together uniquely determine the next state~\cite{Ash65a, Crut10a}.
\end{Def}	
The canonical unifilar presentation---which can have a finite, countably infinite, or uncountably infinite number of recurrent states---will have a distinct
latent state for each 
unique 
probability distribution over possible futures 
that can be induced by some past.  
I.e., we consider observation-induced transitions among 
probability measures $\Pr_{\params}(\vec{X} | w)$, where $\vec{X}$ is an infinite future and $w \in \mathcal{X}^*$ is an observed history
that precedes it.
Any other unifilar presentation is just a redundant version of this.

Suppose that the number of recurrent states in a unifilar model is at most countably infinite.
Then we
find that the information 
rate $\FIrate$ can be calculated directly from any parametrized
unifilar presentation of a process, with matrix elements given by: 
\begin{align}
	(\FIrate)_{m,n} 
	&= 
	- \sum_{s \in \SSet^+}
	\Blackwell(s)
	\sum_{x \in \Abet}
	\Pro(x|s) \,
	\partial_{\theta_m} \partial_{\theta_n}   \ln \Pro(x | s) ~,
	\label{eq:SimpleFIrate}
\end{align}	
or, equivalently
\begin{align}
	(\FIrate)_{m,n} 
	&= 
	 \sum_{s \in \SSet^+}
	\Blackwell(s)
	\sum_{x \in \Abet}
	\frac{\bigl[ \partial_{\theta_m}   \Pro(x | s) \bigr] \bigl[ \partial_{\theta_n}   \Pro(x | s) \bigr] }{ \Pro(x | s) }
	~,
	\label{eq:SimpleFIrate2}
\end{align}	
where $\SSet^+$ is the set of recurrent latent states of the unifilar presentation, $\Blackwell$ is the 
stationary distribution over those states,
and $\Pro(x | s)$ is the probability that the process will next produce the observation $x$ given that it is currently in state $s$.
The smallest such presentation is the 
$\epsilon$-machine of computational mechanics,
 if the topology of the $\epsilon$-machine is invariant to small changes in $\vec{\theta}$~\cite{Shal98a, Crut12a}.~\footnote{At special parameter values, a larger model than the $\epsilon$-machine is needed, e.g., to represent the Perturbed Coin process topology at the parameter setting where it reduces to the Biased Coin process:  the larger Perturbed Coin process topology must still be used.}

The following not only derives this result, but also discovers the convergence properties to this asymptotic information rate.
The myopic information rate 
and its 
accumulated transients also have closed form solutions, given in Secs.~\ref{sec:Transients} and \ref{sec:Excess} respectively, 
that contribute to the exact bound on estimation variance.

If one instead starts with
a parametrized nonunifilar model 
with states $\SSet$,
then
a unifilar 
presentation
can be constructed from
the metadynamic among observation-induced
probability distributions over 
$\SSet$---this is the so-called 
`mixed-state presentation' (MSP)~\cite{Riec18_SSAC1}.
The MSP reproduces the 
same stochastic process as the nonunifilar model,
but is unifilar since (via Bayes' rule)
an observation induces a particular 
updated distribution over those nonunifilar-model states.
Eq.~\eqref{eq:SimpleFIrate} can then be applied 
to the recurrent component of the MSP, since it is 
topologically invariant to generic variations of parameters.
The algorithm to construct the MSP will be presented in Sec.~\ref{sec:HMMs},
where it becomes central to our study, 
since it directly controls the convergence of myopic information.

Sec.~\ref{sec:EntropyVsInfo}
discusses the unexpected but rigorous 
connection between information rates and entropy rates.
Sec.~\ref{sec:InfoVector} 
derives simplified expressions for the information vector that 
will be useful in application.

Different model classes distinguish themselves 
by the complexity of their convergence behavior, as we discuss in Sec.~\ref{sec:Examples}.
Finite Markov-order processes converge exactly to the information rate at finite length.  
Processes with finite unifilar transient structure converge exponentially to the information rate---although with possibly unconstrained ephemeral behavior up to some `index of symmetry collapse' to be discussed, followed by a sum of polynomials times decaying exponentials.
Processes with a finite HMM recurrent structure converge at least exponentially fast to the rate.
Processes with no finite-state HMM presentation can display non-exponential convergence.
We give examples of processes with finite and infinite Markov order, 
and also consider simple examples that are non-stationary and non-ergodic.

Sec.~\ref{sec:MLE}
explores the 
relationship between 
the number of
independent sequences
and sequence length
in determining variance of model-parameter estimates.
We demonstrate that the maximum likelihood estimator (MLE),
which is known to be asymptotically unbiased 
for many independent samples, is also asymptotically unbiased
and efficient with increasing length of a 
single correlated sequence of an ergodic process.

Sec.~\ref{sec:Overparam}
briefly comments on the implications of our results in 
the practically relevant case of overparametrized models.
We then close with Sec.~\ref{sec:Conclusion}.

\section{Metadynamic among estimation states}
\label{sec:Transients}

Expanding Eq.~\eqref{eq:MyopicInfoExpression},
elements of the myopic information rate can be written explicitly
as 
\begin{align}
	(\dF)_{m,n}
	&= - \!\! \sum_{w \in \mathcal{X}^{L-1}}
	\sum_{x \in \mathcal{X}}
	\Pr_{\params}(X_{1:L-1} = w )
	\Pr_{\params}(X_L = x | X_{1:L-1} = w ) \,
	 \partial_{\theta_m} \partial_{\theta_n}   \ln \Pr_{\params}(X_L = x | X_{1:L-1} = w)  
	 \label{eq:dF_explicit}
\end{align}	
for $L \geq 2$.
This formulation suggests that we need exponentially diverging resources to calculate the myopic information rate at longer lengths 
since the number $|\mathcal{X}|^{L-1}$ of possible histories grows exponentially.
However, not all histories need to be distinguished:
For the purpose of prediction and estimation, we 
can cluster those histories that yield the same 
future predictions,
as done routinely in computational mechanics~\cite{Uppe97a, Shal98a,  Crut12a}.
This significantly reduces the resources required to calculate the myopic information rate.

To progress, we consider
the 
equivalence classes
of 
arbitrary-length histories $w \in \mathcal{X}^*$
that induce the same conditional probability distributions over futures.
\begin{Def}
Two histories 
$w$ and $w'$
(of arbitrary lengths 
$\ell$ and $\ell'$)
belong to the same 
 \emph{estimation state}
$\es(w) = \es(w') \in \esSet$
if 
$ \Pr_{\params'}( X_{\ell+1:\ell+k} | X_{1:\ell} = w) = 
\Pr_{\params'}( X_{\ell'+1:\ell'+k} | X_{1:\ell'} = w') \; \text{ for all } \params'$ 
sufficiently close to $\params$ and for all $k \geq 1$.
\end{Def}
The set of 
estimation states
$\esSet$
thus partitions the set of all histories $\mathcal{X}^*$,
and we have invoked the 
\emph{estimation-equivalence function} 
$\es \colon  \mathcal{X}^* \to \esSet$
that 
maps an observed history of any length to 
its 
estimation state.
We will also consider the empty history $w =  \varnothing$ when no observations have been made, which belongs to the 
estimation state
$\es(\varnothing) = \bigr\{ w \in \mathcal{X}^* : \Pr_{\params}( X_{|w|+1:|w|+k} | X_{1:|w|} = w) = \Pr_{\params}( X_{1:k} ) \text{ for all } \params'$ 
sufficiently close to $\params$ and for all $k \geq 1 \bigr\}$.

These estimation states
$s \in \esSet$
are closely related to the `process states' of computational mechanics~\cite{Uppe97a}, the recurrent states of which are called `causal states'~\cite{Crut12a}.
However, 
there is an important albeit subtle difference:
the estimation-equivalence relation 
requires `causal equivalence' of two words throughout a neighborhood of $\params$---not just when the process is evaluated at $\params$.

Since the observed sequence is a random variable,
the estimation state
$\ES_L$ after 
$L-1$ observations is also a random variable.
It is fruitful to
construct a vector space
with
an orthonormal basis $\{ \ket{\Wstate[s]} \}_{s \in \esSet}$
corresponding to
the set of estimation states.
We consider a set of linear operators on this vector space $\{ W^{(x)} \}_{x \in \mathcal{X}}$
with transition 
elements 
\begin{align}
W^{(x)}_{s,s'} 
&= 
\Pr_{\params}(\ES_{t+1} = s', X_t = x | \ES_t = s) \\
&= 
\Pr_{\params}( X_t = x | \ES_t = s) \, \delta_{wx \in s'} & \text{for any } w \in s 
~.
\end{align}
Here, $wx$ is a concatenated word, and $ \delta_{wx \in s'}$ is an indicator function equal to 1 if $wx \in s'$ and equal to zero otherwise.
Notice that $W^{(x)}$ is independent of $t$
since conditioning on an 
estimation state renders time insignificant.
Constructively,
in Sec.~\ref{sec:HMMs},
we will show how to calculate these unifilar transition elements $W^{(x)}_{s,s'}$ directly from any HMM presentation of the process.

These substochastic labeled transition matrices sum to the 
net stochastic transition matrix 
$W = \sum_{x} W^{(x)}$
among 
estimation states.
If the metadynamic has a single attractor, then
its right eigenstate of all ones $\Wones = W  \Wones = [1 , 1, \dots , 1]^\top$ 
indicates conservation of probability,
while
its left eigenstate $\Wpi= \Wpi  W = \frac{ \Wpi }{\langle \stationary  \Wones }$
is the stationary distribution over 
recurrent
estimation states.
However, in the case of nonergodic processes,
there can be multiple attractors.

The sum over all individual histories in Eq.~\eqref{eq:dF_explicit}
can now be replaced by a sum over equivalence classes of histories
since they yield the same probabilities $\Pr_{\params}(X_L = x | X_{1:L-1} = w ) = \Pr_{\params} \bigl( X_L = x |  \ES_L = \es(w) \bigr)$. 
 Before any observations, 
the 
estimation state is 
$\es(\varnothing)$,
so $\Pr_{\params}(\ES_1) = \delta_{\ES_1, \es(\varnothing)} = \StartMS  \Wstate[\ES_1] \rangle$
where $\bra{\Wstate[s]} = [0, \dots , 0,  1,  0, \dots , 0]$ can be thought of as
a row vector of all zeros except a one at the position for estimation state $s$, while $\ket{\Wstate[s]} = \bra{\Wstate[s]}^\top$.
After $L-1$ observations,
$\Pr_{\params}(\ES_L = s) = \StartMS  W^{\ell-1} \ket{ \Wstate[s] }$.

As shown explicitly in App.~\ref{sec:ddev},
the myopic information rate now becomes
\begin{align}
	(\dF)_{m,n}
&= - \!\! \sum_{s \in \esSet}
	\sum_{x \in \mathcal{X}}
	\Pr_{\params}(\ES_L = s )
	\Pr_{\params}(X_L = x | \ES_L = s) \,
	\partial_{\theta_m} \partial_{\theta_n}   \ln \Pr_{\params}(X_L = x | \ES_L = s)  
\end{align}	
for any integer $L \geq 1$, 
which can profitably be rewritten as 
\begin{align}
	(\dF)_{m,n}
&= \StartMS W^{L-1} \ket{ f_{m,n} } ~,
	\label{eq:MyopicInfoViaWL}
\end{align}	
where
we have introduced the $L$-independent \emph{information vector}
\begin{align}
	\ket{ f_{m,n} } 
	&
\coloneqq 
- \!\! \sum_{s \in \esSet}
\sum_{x \in \mathcal{X}}
\ket{\Wstate[s]}
\Pro( x | s) \,
\partial_{\theta_m} \partial_{\theta_n}   \ln \Pro( x |  s)  
	~.
	\label{eq:fmn_ket_def}
\end{align}	

For explicit calculations, it may be useful to recognize two alternative expressions for the information vector:
\begin{align}
	\ket{ f_{m,n} } 
	&=
	\sum_{s \in \esSet}
	\sum_{x \in \mathcal{X}}
	\ket{\Wstate[s]}
	\Pro( x |  s) 
	\bigl[ \partial_{\theta_m} \ln	\Pro(x | s) \bigr]  \bigl[ \partial_{\theta_n} \ln	\Pro( x |  s) \bigr]  
		\label{eq:fmn_ket_def2}
\end{align}	
and
\begin{align}
	\ket{ f_{m,n} } 
	&=
	\sum_{s \in \esSet}
	\sum_{x \in \mathcal{X}}
	\ket{\Wstate[s]}
	\frac{\bigl[ \partial_{\theta_m} 	\Pro(x | s) \bigr]  \bigl[ \partial_{\theta_n} 	\Pro( x |  s) \bigr]  }{ \Pro( x |  s) }
	\label{eq:fmn_ket_def3}
	~.
\end{align}	
In the explicit calculations for our later examples, the latter version, Eq.~\eqref{eq:fmn_ket_def3}, will be especially useful. 

Eq.~\eqref{eq:MyopicInfoViaWL}
is a powerful result, because it says that the myopic information rate depends on the observation length only via powers of the linear operator $W$.
A spectral decomposition of $W^L$ thus reveals all behavior of convergence to the asymptotic information rate.
However $W$ is not typically a normal operator---i.e., $W^\dagger W \neq W W^\dagger$---so the spectral theorem for normal operators does not apply.  
(Note that non-normality is already implied by $\Wpi \neq \Wones^\top$.)  Moreover, $W$ is often nondiagonalizable due to an eigenvalue of zero that enables non-exponential ephemeral behavior.
Nevertheless, a spectral decomposition of powers of the non-diagonalizable operator is both possible and illuminating, using the techniques developed in Refs.~\cite{Riec18_Beyond, Riec18_SSAC2}.

Recall the definition of $W$'s spectrum, 
$\Lambda_W \coloneqq \{ \lambda \in \mathbb{C} : W - \lambda I \text{ is not invertible} \}$, 
where $I$ denotes the identity operator.
The
\emph{index} $\nu_\lambda$ of the eigenvalue $\lambda$
is the  size of the largest Jordan
block associated with $\lambda$. 
(For a diagonalizable operator, the index of all eigenvalues is one. The index must be larger than one if the algebraic multiplicity is larger than the geometric multiplicity.)

Let's assume that 
$W$'s spectrum $\Lambda_W$ 
consists of isolated eigenvalues---which
will generically be the case for such a transition operator, and will certainly be the case for any finite matrix approximation of $W$.
Then we can 
employ the general spectral decomposition
of $W^{L-1}$ derived in Ref.~\cite{Riec18_Beyond} for nonnormal and nondiagonalizable operators:
\begin{align}
	W^{L-1} &= 
	\Bigl[ \sum_{k=0}^{\nu_0 - 1} \delta_{L-1, k} W_{0,k} \Bigr]
	\! + \!\!\!\! \sum_{\lambda \in \Lambda_W \setminus \{ 0 \} } \sum_{k=0}^{\nu_\lambda - 1} 
	\binom{L-1}{k}
	\lambda^{L - k -1}
	W_{\lambda, k}
	,
	\label{eq:SpectralDecompOfPowersOfW}
\end{align}
where $\binom{L-1}{k}$ is the generalized binomial coefficient:
\begin{align}
	\binom{L-1}{k} &= \frac{1}{k!} \prod_{\ell=1}^k (L - \ell)
	~,
\end{align}
with $\binom{L-1}{0} = 1$.
The spectral companion operator is 
\begin{align}
W_{\lambda, k} = W_\lambda (W- \lambda I
)^k ~.
\end{align} 
The spectral projection operator $W_{\lambda}$ associated with
eigenvalue $\lambda$ can be defined as the residue of the resolvent $(z I - W)^{-1}$ as
$z \to \lambda$:
\begin{align}
	W_{\lambda} = \tfrac{1}{2 \pi i} \oint_{\gamma_\lambda} \bigl( zI - W \bigr)^{-1} \, dz
	~,
\end{align}
where $z \in \mathbb{C}$ and $\gamma_\lambda$ is a small counterclockwise contour
around the eigenvalue $\lambda$. 

Alternatively, the spectral projection
operator
$W_\lambda$
can be constructed from all left eigenvectors, generalized left
eigenvectors, right eigenvectors, and generalized right eigenvectors
associated with $\lambda$. The construction is given explicitly in
Ref.~\cite{Riec18_Beyond}.
When the algebraic multiplicity of an eigenvalue is one,
its spectral projection operator is simply
$W_\lambda = \frac{\ket{\lambda} \bra{\lambda}}{\braket{\lambda| \lambda}}$, in terms 
of its right eigenstate $W \ket{\lambda} = \lambda \ket{\lambda}$ and its left eigenstate
$\bra{\lambda} W = \lambda \bra{\lambda}$.

As a transition matrix,
all of $W$'s eigenvalues lie on or within the unit circle in the complex plane.  An eigenvalue $\lambda=1$ is associated with the stationary distribution.  Other eigenvalues on the unit circle correspond to deterministic periodicities, while eigenvalues within the unit circle correspond to modes of relaxation since $|\lambda| < 1$. 

Substituting Eq.~\eqref{eq:SpectralDecompOfPowersOfW} into
Eq.~\eqref{eq:MyopicInfoViaWL} yields:
\begin{align}
		(\dF)_{m,n}
	&=  
	\Bigl[ \sum_{k=0}^{\nu_0 - 1} \delta_{L-1, k}  \StartMS  \, W_{0,k}  \,  \ket{ f_{m,n} }  \Bigr] 
+ \sum_{\lambda \in \Lambda_W \setminus \{ 0 \} } \sum_{k=0}^{\nu_\lambda - 1} 
	\binom{L-1}{k}
	\lambda^{L - k -1}
	\StartMS  \, W_{\lambda,k}  \,  \ket{ f_{m,n} } 
	~,
\end{align}
for all $L \geq 1$.

It is significant that the zero eigenvalue contributes a qualitatively distinct
ephemeral behavior to the myopic information rate while $L  \leq \nu_0$. 
This is discussed further in Sec.~\ref{sec:SymmetryCollapse}.
All other
eigenmodes contribute products of polynomials times decaying exponentials in
$L$. 
Typically $W$ will be `almost diagonalizable'---i.e.,
diagonalizable on all eigenspaces except for possibly the zero-egeinvalue subspace.
When $W$ is diagonalizable and invertible, the myopic information is simply a sum of
decaying exponentials.

If there are no periodic oscillations,
then the information rate is 
\begin{align}
	(\FIrate)_{m,n} 
	&= \lim_{L \to \infty}
		(\dF)_{m,n} 
=  
	\StartMS  \, W_{1}  \,  \ket{ f_{m,n} } 
	~. 
\end{align}	
When there is a single attractor,
$W_1 = \Wones \! \Wpi$,
and so
\begin{align}
	(\FIrate)_{m,n} 
&= \Wpi  f_{m,n} \rangle ~.
	\label{eq:ConciseInformationRateFormula}
\end{align}	

By nature of the estimation-equivalence relation, 
the estimation metadynamic $\{ W^{(x)} \}_x$ 
is the minimal unifilar model valid throughout a small neighborhood of $\params$.
Any non-minimal unifilar model is thus a fine graining of histories that splits its states without affecting those states' overlap with $\ket{f_{m,n}}$.
If the row vector $\stationary$ represents the stationary distribution over the states of a unifilar model,
and if the column vector $\vec{f}_{m,n}$ represents 
the information vector Eq.~\eqref{eq:fmn_ket_def} restricted to these states,
then we can concisely write the elements of the asymptotic Fisher information rate as 
\begin{align}
	(\FIrate)_{m,n} 
	&= \stationary  \vec{f}_{m,n} ~.
	\label{eq:ConciseInformationRateFormula2}
\end{align}	
Writing 
Eq.~\eqref{eq:ConciseInformationRateFormula2} out explicitly 
thus yields one of our main results, already stated as 
Eq.~\eqref{eq:SimpleFIrate}.

\section{Excess information}
\label{sec:Excess}

The excess information,
present in the myopic information
beyond its asymptotic rate,
can be accumulated analytically too 
with the help of 
the explicit metadynamic among estimation states.  In particular, we find 
\begin{align}
	\EFI 
	&= \sum_{\ell=1}^L	(\dF[\ell] - \FIrate) 
= \sum_{\ell=1}^L
	\Bigl[  \StartMS W^{L-1} \ket{ f } 
	-  \StartMS W_1 \ket{ f } 
	\Bigr] \\
	& = 
	\StartMS \Bigl[ \sum_{\ell=1}^L (W - W_1 )^{L-1} \Bigr] \ket{ f } 
	-  \FIrate \\
	& = 
	\StartMS (I-Q)^{-1} (I-Q^L) \ket{ f } 
	-  \FIrate ~,
\end{align}	
where $Q \coloneqq W -  W_1$ is invertible by construction.  Recall that, if there is a single attractor, then $W_1 = \Wones \Wpi$.
For convenience, we have used a
`ket' $\ket{f}$ notation, 
which is really a matrix of column vectors $ \ket{ f_{m,n} } $---one for each element of the information matrix.
Since $\lim_{L \to \infty} Q^L = 0$ 
in the typical case without deterministic periodicities,
the asymptotic excess information becomes
\begin{align}
	\boldsymbol{\epsilon} 
	= 
	\StartMS (I-Q)^{-1} \ket{ f } 
	-  \FIrate 
	\label{eq:SpectralEFI}
\end{align}	
whenever a finite limit exists.
Elementwise, this can be written as 
$	( \boldsymbol{\epsilon} )_{m,n}
= 
\StartMS (I-Q)^{-1} \ket{ f_{m,n} } 
- ( \FIrate )_{m,n} $.

\section{Hidden Markov models and belief states}
\label{sec:HMMs}

The preceding development has been intentionally general, without reference to a generative mechanism for the stochastic process.
However, if a parametrized HMM representation of the process is known---whether unifilar or nonunifilar, and whether finite state or infinite state---then the transition matrices of this HMM can be used to directly calculate the transition probabilities between estimation states.  Meanwhile, the estimation states themselves are generated via the observation-induced belief states (or `mixed states') about the latent states of the generating HMM. 
The following construction is thus known as the `mixed state presentation' (MSP) in computational mechanics~\cite{Uppe97a, Riec18_SSAC1, Jurg21_Shannon}.

A parametrized
Mealy hidden Markov model (HMM)
presentation of a parametrized process
can be specified by the 3-tuple 
$\mathcal{M} = 
\bigl( 
\mathcal{X}, \,
( T^{(x)} )_{x \in \mathcal{X}}, \,
\boldsymbol{\mu}
\bigr)$.
In this specification,
the labeled transition matrices 
$\bigl( T^{(x)} \bigr)_{x \in \mathcal{X}}$
indicate transition probabilities
$T^{(x)}_{s,s'} = \Pr_{\params}(\St_{t+1} = s', X_t = x | \St_t = s)$
between hidden states $s, s' \in \SSet$ of the model, 
while $\boldsymbol{\mu}$ is the initial distribution over latent states.
Our simple notation for transition matrices $T^{(x)}$ and initial distribution $\boldsymbol{\mu}$
keeps parameter dependence implicit.

The labeled transition matrices sum to the net stochastic transition matrix $T = \sum_{x \in \mathcal{X}} T^{(x)}$ 
among hidden states.
Its right eigenstate of all ones $\Tones = T \Tones = [1 , 1, \dots , 1]^\top$ 
indicates conservation of probability,
while
its left eigenstate $\Tpi = \Tpi  T = \Tpi  / \Tpi \Tones$
is the stationary distribution over hidden states.
A stationary process can be specified by setting 
$\boldsymbol{\mu} = \stationary_T$.

The probability of any word $w = x_1 \dots x_\ell$
is given by 
\begin{align}
\Pr_{\params}(X_{1:\ell} = w) = \boldsymbol{\mu}  T^{(w)} \Tones = \boldsymbol{\mu} T^{(x_1)} \cdots T^{(x_\ell)} \Tones ~.
\end{align}
The probability distribution over an arbitrary length-$k$ future, given some history is thus given by
$\Pr_{\params}(X_{\ell+1:\ell+k} | X_{1:\ell} = w) = \boldsymbol{\mu}  T^{(w)} T^{(X_{\ell+1:\ell+k})} \Tones /  \boldsymbol{\mu}  T^{(w)} \Tones= 
\frac{ \boldsymbol{\mu} T^{(w)}  }{  \boldsymbol{\mu}  T^{(w)} \Tones }
T^{(X_{0:k})} \Tones$.
Notice that this conditional probability over the future depends on the 
history only via the 
probability vector or `mixed state' 
\begin{align}
\eta_{\params}^{(w)} = 
\frac{ \boldsymbol{\mu} T^{(w)}  }{  \boldsymbol{\mu}  T^{(w)} \Tones } 
\end{align}
that it induces.
Thus, any history with the same probability vector induces the same probability distribution over futures.

Mixed states are thus closely related but subtly different from estimation states.
Non-minimal models can have redundancies, where different distributions over model states can yield the same probability distribution over futures~\cite{Uppe97a}.
The 
estimation-equivalence class of a particular history
can
generally
be seen 
as the set of all histories that induce the same probability distribution over futures:
\begin{align}
\es(w) 
= 
\bigl\{ w' \in \mathcal{X}^* : \eta_{\params}^{(w)} T^{(w'')} \Tones = \eta_{\params}^{(w')} T^{(w'')} \Tones \text{ for all } w'' \in \mathcal{X}^*  \bigr\} ~.
\end{align}	
This allows 
two different mixed states 
$\eta_{\params}^{(w)} \neq \eta_{\params}^{(w')}$
to correspond to the same estimation state
$\es(w) = \es(w') $
when
they produce no observable differences in probabilities, such that their differences 
$\eta_{\params}^{(w)} - \eta_{\params}^{(w')}$
are orthogonal to the `future space'
$\mathcal{F} = \text{span} \{ T^{(w)} \Tones : w \in \mathcal{X}^* \}$.
For minimal models---those without such redundancy---estimation states are simply 
isomorphic to
the parametrized mixed states:
$\es(w) = \bigl\{ w' \in \mathcal{X}^* : \eta_{\params}^{(w)} = \eta_{\params}^{(w')}  \bigr\}$.

In either case, mixed states allow direct calculation of the estimation-state transition probabilities appearing in Eq.~\eqref{eq:fmn_ket_def}:
\begin{align}
\Pro( x |  s) 
	&=
	\eta_{\params}^{(w)} T^{(x)} \Tones &\text{for any } w \in s ~.
\end{align}	

From the mixed state $\eta_{\params}^{(w)} $,
upon observing $x$
there is a unifilar transition to the new mixed state 
$\eta_{\params}^{(wx)} $.
Each of these mixed states correspond either to 
an estimation state or a fine graining of an estimation state.

We leverage this 
parametrized mixed-state construction 
to determine both the 
information rate and excess information
in the following examples.
However, in App.~\ref{sec:BlackwellAnalog} we note that
this parameter dependence---the fact that $\partial_{\theta_n} \eta_{\params}$ 
can be
nonzero---complicates attempts to 
usefully express 
the information rate via a Lebesgue integral over the mixed-state simplex, which would be 
analogous to the Blackwell expression for the entropy rate of a process~\cite{Blac57b, Jurg21_Shannon}.

\section{Entropy vs.\ information, and exponential convergence to Fisher-information rate}
\label{sec:EntropyVsInfo}

Our closed-form expressions for the myopic information rate and excess information reveal
an exact analogy with myopic entropy rate and excess entropy.
The myopic entropy rate
$h_\ell$
quantifies the apparent entropy rate of the best Markov-order-$\ell$
approximation of the process~\cite{Crut01a}.
The myopic entropy thus
reveals interdependencies 
and hidden correlations among random variables,
whether they are near, distant, or only correlated through very high orders of correlation~\cite{Riec21_Fraudulent}. 
Recall that the Shannon entropy of a random variable $X \sim \Pr(X)$
is $\H[X] = - \sum_x \Pr(X=x) \log \Pr(X=x)$, 
while the conditional entropy of $X$ given $Y$, if $X$ and $Y$ are 
jointly distributed as $\Pr(X,Y)$, is 
$\H[X|Y] = \H[X,Y] - \H[Y]$
~\cite{Cove12}.
The base of the logarithm is arbitrary but should be specified for units of bits, nats, etc.
In particular, the myopic entropy rate can be expressed as~\cite{Crut16_Exact}
\begin{align}
	h_\ell 
	&= \H[X_\ell | X_{1:\ell-1} ] \\
	&= \StartMS W^{\ell-1} \ket{\H(W^{\mathcal{X}})} ~,
	\label{eq:MyopicEntropy}
\end{align}	
where $ \ket{\H(W^{\mathcal{X}})} \coloneqq - \sum_{s \in \esSet} \ket{s} \sum_{x \in \mathcal{X}} \Pro(x | s) \log \Pro(x | s)$,
yielding the asymptotic Shannon entropy rate
$h = \lim_{\ell \to \infty} h_\ell = \Wpi \H(W^{\mathcal{X}}) \rangle = \stationary_T \vec{\H}(T^{\mathcal{X}})  $ for any unifilar model.
Eq.~\eqref{eq:MyopicEntropy} 
has a clear and rigorous analogy with 
Eq.~\eqref{eq:MyopicInfoViaWL}.
The 
mutual information between past and future 
of a stationary stochastic process,
also known as the
`excess entropy', can be expressed as~\cite{Riec18_SSAC2}
\begin{align}
\textbf{E}
&= \lim_{L \to \infty} \text{I}[X_{-L+1:0} ; X_{1:L}]
	= \sum_{\ell = 1}^\infty (h_\ell - h) \\
	&= \StartMS (I-Q)^{-1} \ket{\H(W^{\mathcal{X}})}  - h ~.
	\label{eq:EE}
\end{align}	
Again,
Eq.~\eqref{eq:EE} 
has a clear and rigorous analogy with 
Eq.~\eqref{eq:SpectralEFI}.

The myopic Fisher information quantifies how the variance of parameter estimation scales with longer observed sequence length 
when learning an \emph{unknown} process.
In contrast,
myopic Shannon entropy quantifies residual uncertainty 
as a function of observed sequence length
while synchronizing to a \emph{known} stochastic process.
Despite this stark difference of whether the process is known or unknown,
we find that the mixed-state metadynamic
controls the behavior of both synchronization and learning.

Since we have shown that myopic Shannon-entropy rate and myopic Fisher-information rate have the same observation-length dependence,
$\StartMS W^{\ell-1} \ket{ \cdot }$,
we can conclude that exponential convergence of the former~\cite{Trav14a} 
implies exponential convergence of the latter 
(given some gentle assumptions on the behavior of the elements of the vectors $\ket{f_{m,n}}$ and $\ket{\H(W^{\mathcal{X}})}$ ).
With this correspondence, 
we can provide an example process with infinite excess information simply by importing an example constructed for infinite excess entropy~\cite{Trav11b}, and imparting it with some parametrization.

For 
any process,
the convergence 
of both 
information and entropy
is determined by the eigenvalues of the MSP transient structure with magnitude between zero and one, according to Eq.~\eqref{eq:SpectralDecompOfPowersOfW}.
For stationary stochastic processes, 
we observe that only those eigenvalues associated with the MSP's transient structure contribute to the convergence:
If $T$ denotes the transition matrix among recurrent estimation states, such that the full estimation-state metadynamic has the block representation
$W = \begin{bmatrix} A & B \\ \boldsymbol{0} & T \end{bmatrix}$,
then we find that 
\begin{align} 
 \StartMS W_\lambda = \vec{0} \text{ for all } \lambda \in \bigl\{ \xi \in \Lambda_T \setminus \{ 1 \} : \xi \notin \Lambda_A  \bigr\}
\end{align}
for any stationary process.
This has important and direct implications for the 
convergence behavior of
either learning or synchronization.

Zero eigenvalues of $W$ determine unique ephemeral structure.
These zero eigenvalues are all that is present in $W$'s transients in the finite-Markov case, but can coexist with other transient modes
to yield the
`symmetry-collapse index'~\cite{Riec18_SSAC2} of 
infinite-Markov-order processes.

\begin{table}[t]
	\centering
\includegraphics[width=0.875\columnwidth]{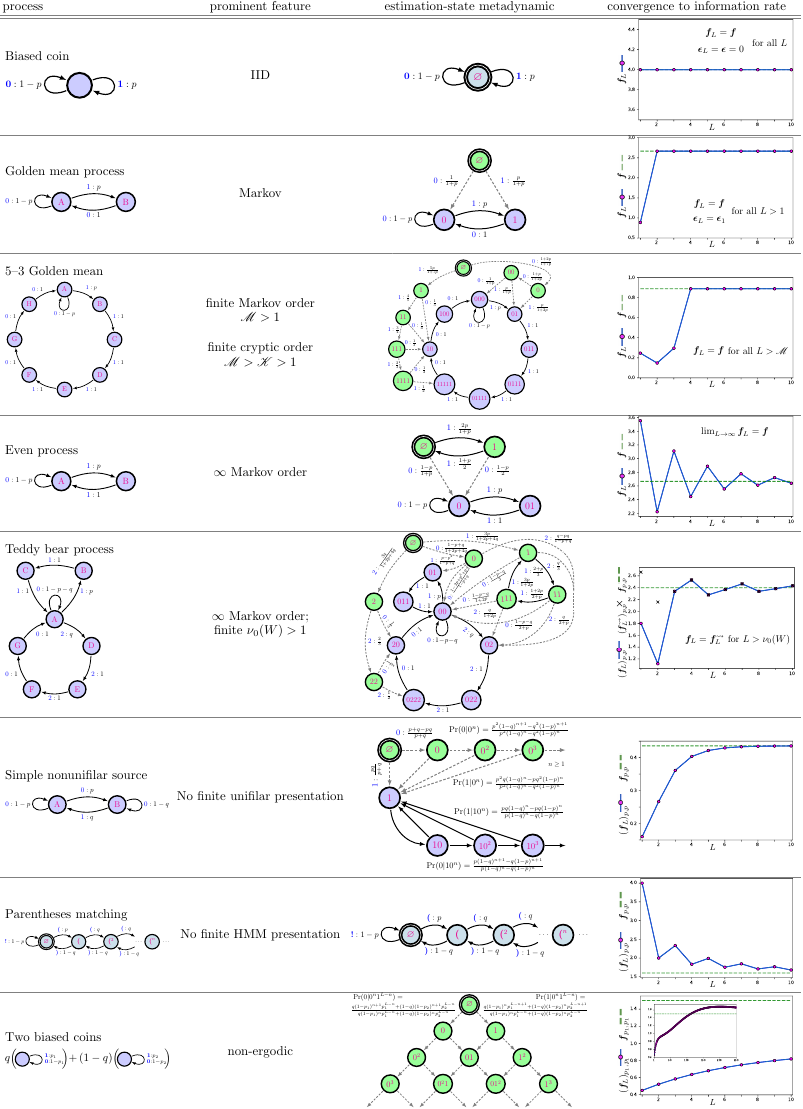}
	\caption{Convergence to Fisher information rate 
		for stochastic processes of approximately increasing complexity.
Qualitatively new features of convergence are exhibited at each row of this table.
		Estimation states are labeled by the shortest word that induces them.  Transient states are colored green.  Dashed gray arrows indicate that the 
		state of origin will never be revisited.
In the rightmost column, convergence to the asymptotic information rate is evaluated at $p=1/2$, $q=1/3$, $p_1 = 1/3$, and $p_2 = 1/2$.
	}
	\label{tab:ProcessTable}
\end{table}

\section{Simplifications of the information vector}
\label{sec:InfoVector}

With Eq.~\eqref{eq:SimpleFIrate2}
the asymptotic information rate is simple to calculate
once we have a unifilar representation of the process---we then just need the stationary distribution and the information vector.
The information vector is straightforward to construct from
the unifilar automaton (or from the corresponding labeled transition matrices) 
in any case via Eq.~\eqref{eq:fmn_ket_def3},
but 
the following tricks 
make the construction even easier.

In our examples to follow, 
transition probabilities out of each state are always 
of the form $\Pro(x|s) = \tfrac{u_x}{g_x}$,
where $u_x$ and $g_x$ are functions  of the parameters $\params$.
Applying the `quotient rule' of elementary calculus to Eq.~\eqref{eq:fmn_ket_def3},
we find it useful to express
\begin{align}
	\braket{s | f_{m,n}} =   \sum_{x \in \mathcal{X}} \frac{\Bigl[ \bigl( \partial_{\theta_m} u_x \bigr) g_x -  u_x \bigl( \partial_{\theta_m} g_x \bigr)  \Bigr] \Bigl[ \bigl( \partial_{\theta_n} u_x \bigr) g_x -  u_x \bigl( \partial_{\theta_n} g_x \bigr)  \Bigr] }{ g_x^3 u_x } ~.
	\label{eq:QuotientRule_Simplification0}
\end{align}

For binary alphabets $\mathcal{X} = \{ 0, 1 \}$, we find that 
\begin{align}
	\ket{f_{m,n}} = \sum_s \ket{s}  \frac{ \bigl[ \partial_{\theta_m} \Pro(0|s) \bigr]  \bigl[\partial_{\theta_n} \Pro(0|s) \bigr] }{ \Pro(0|s) \Pro(1|s) } ~,
\end{align}	
or equivalently
\begin{align}
	\ket{f_{m,n}} = \sum_s \ket{s}  \frac{\bigl[ \partial_{\theta_m} \Pro(1|s) \bigr] \bigl[ \partial_{\theta_n} \Pro(1|s) \bigr] }{ \Pro(0|s) \Pro(1|s) } ~.
	\label{eq:BinarySimplification1}
\end{align}	
If the transitions out of an estimation state are given as a ratio $\Pro(x|s) = \tfrac{u}{g}$ for either $x=0$ or $x=1$,
where $u$ and $g$ are both functions of the parameters $\params$,
then the `quotient rule' of elementary calculus can again be employed for binary processes to eventually yield
\begin{align}
	\braket{s | f_{m,n}} =   \frac{\Bigl[ \bigl( \partial_{\theta_m} u \bigr) g -  u \bigl( \partial_{\theta_m} g \bigr)  \Bigr] \Bigl[ \bigl( \partial_{\theta_n} u \bigr) g -  u \bigl( \partial_{\theta_n} g \bigr)  \Bigr] }{ g^2 u(g-u) } ~.
	\label{eq:QuotientRule_Simplification1}
\end{align}

Even more simply, if there is a single parameter $\params = \theta$, then
\begin{align}
	\ket{f} &= \sum_s \ket{s}  \frac{\bigl[ \partial_{\theta} \text{P}_{\theta}(1|s) \bigr]^2 }{ \text{P}_\theta(0|s) \text{P}_{\theta}(1|s) } 
\label{eq:BinarySimplification2}
\end{align}	
and 
\begin{align}
	\braket{s | f}
	=
	\frac{\bigl[ ( \partial_{\theta} u ) g -  u ( \partial_{\theta} g )  \bigr]^2 }{ g^2 u(g-u) } 
	~.
	\label{eq:BinarySimplification3}
\end{align}	
To derive these expressions,
we 
start with Eq.~\eqref{eq:fmn_ket_def3} and
utilize the fact that $\Pro(1|s) = 1 - \Pro(0|s)$ for binary processes.
We leverage these simplification to obtain closed-form expressions in many of the following examples.
When implemented numerically, the information vector is easy to compute as a function of the mixed-state labeled transition matrices $(W^{(x)})_{x \in \mathcal{X}}$ and their element-wise partial derivatives, via Eq.~\eqref{eq:fmn_ket_def3}, for any alphabet size and any number of parameters.

\section{A partially ordered hierarchy of complexity}
\label{sec:Examples}

We 
now embark of a brief tour through increasingly complex processes.
Our goal of this section is to highlight some of the
qualitatively distinct features that can arise
in the parameter estimation of complex processes---as we move from memoryless, to Markov, to finite-order Markov, to infinite-order Markov, and then to yet more sophisticated model classes.
To facilitate this, 
we will step through the processes represented in each row of Table \ref{tab:ProcessTable}, and will comment on 
the implications for 
inferring their structure from data.
All of the processes we will examine are stationary,
except for the Parentheses Matching process which 
has a nonstationary initial distribution over its recurrent states.
The automata in Table \ref{tab:ProcessTable} and in the following,
with directed edges from state $s$ to $s'$,
are labeled as
 ``{\color{blue}$x$} $: \, \Pr_{\params}(\St_{t+1} = s' , \, X_t = x | \St_t = s )$''
by 
the observable output $x \in \mathcal{X}$ and the $x$-dependent transition probability between those latent states.

The following examples illustrate how different complexity classes induce different convergence behaviors towards the asymptotic information rate.
We encourage the reader to study and work through
App.~\ref{sec:Even},
which can serve as a useful tutorial 
on how to apply our general 
theoretical results to
calculate closed-form expressions for both the 
asymptotic and
myopic information rates in
these examples.
This partially ordered hierarchy is non-exhaustive but begins to indicate some of the intricacies of inference.

\subsection{IID}

Independent and identically distributed (IID) processes
require no memory of their past behavior to produce their future behavior.  These are the simplest stochastic processes, where every observation is simply drawn from the same distribution.

IID processes have a single estimation state.
The Fisher information thus scales trivially 
and exactly linearly for these processes
as longer sequences are observed:  
$\FI = L \FIrate$.
In other words, 
the myopic information rate is exactly the asymptotic information rate at every length $\FIrate_L = \FIrate$, and 
the excess information is zero at every length $\EFI = 0$.
The rate of learning does not change with longer observations.

The simple 
Biased Coin process,
shown in the first row of Fig.~\ref{tab:ProcessTable},
exemplifies this IID class of processes.
From Eq.~\eqref{eq:BinarySimplification2}, 
we immediately see that its Fisher information rate is 
$\FIrate = \tfrac{1}{p(1-p)}$.

\subsection{Markov}

The future of a Markov process is conditionally independent of the past output, given the most recent output:
$\Pr_{\params}(X_{\ell+1:L} | X_{1:\ell}) = \Pr_{\params}(X_{\ell+1:L} | X_{\ell}) $.

The estimation-state metadynamic of a Markov process has a single transient state 
that is completely flushed of probability after one time step.  
This corresponds to an eigenvalue of zero in the 
estimation-state-to-state transition dynamic.
In contrast, the number of recurrent estimation states is upper bounded by 
the size of the observable alphabet $|\mathcal{X}|$
for a Markov process.
The myopic information rate converges to the asymptotic information rate exactly at a block length of two: $\FIrate = \FIrate_2$. 
For Markov processes,
the net Fisher information provided by sequences of any length $L \geq 1$ is thus
$\FI = L \FIrate +  \EFI[1]$, 
where $\EFI[1] =  \FIrate_1 -  \FIrate$.

The Golden Mean process 
in the second row of Table \ref{tab:ProcessTable}
is a paradigmatic example of a Markov process.  With a binary alphabet, its only topological constraint is that it can't produce consecutive `0's.
After seeing either a `0' or a `1',
an observer is fully synchronized to the process, which explains the simple transition structure of the estimation-state metadynamic.

\subsection{Finite Markov order}

As first pointed out by Ref.~\cite{Radaelli22_Fisher},
the excess information saturates at the Markov order
$\mathscr{M}$ of the process: $\boldsymbol{\epsilon} = \EFI[\mathscr{M}]$.
This implies that the 
myopic information rate exactly converges
to the asymptotic information rate 
at a block length of $\mathscr{M}+1$.
Thus, 
the net Fisher information provided by sequences of length $L \geq \mathscr{M}$ is 
$\FI = L \FIrate +  \EFI[\mathscr{M}]$.
From our new perspective, we can now see what this means for the structure of the 
metadynamic among estimation states.

For processes of finite Markov order $\mathscr{M}$,
the MSP has a completely ephemeral tree-like transient structure of depth $\mathscr{M}$.
This corresponds to eigenvalues of zero,
and nondiagonalizability of the zero eigenspace
with index $\nu_0 = \mathscr{M}$.

The $5$-$3$ Golden Mean process in the third row of Table \ref{tab:ProcessTable} provides an example of a process with finite Markov order $\mathscr{M}=5$ and finite cryptic order $\co=3$.

The \emph{cryptic order} 
\begin{align}
	\co \coloneqq \min \{ k \in \mathbb{N}_{0} : \H[\ES_{k+1} | X_{1:\infty}] = 0 \}
\end{align}	
of a process 
is always upper bounded by the Markov order:
$\co \leq \mathscr{M}$.
It
tells us how far into the past we would have to look
to resolve the current estimation state, given the entire future.
At first encounter, cryptic order
may seem like a strange and indeed cryptic quantity.
Nevertheless, it plays a pivotal role in determining the 
causal irreversibility of processes~\cite{Crut08a}
and the extent to which quantum memory can compress the generator of a classical stochastic process~\cite{Maho15a, Riec16_Minimized, Bind18}.
Here we will see that the cryptic order
also plays an unexpected role in the fundamental limits of learning a process.

\begin{figure}
	\includegraphics[width=0.28\columnwidth]{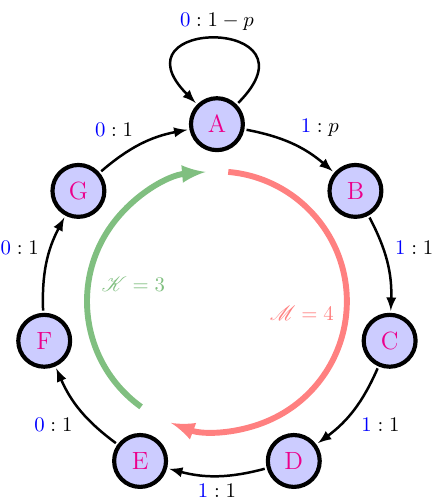}
	\caption{The $\mathscr{M}$-$\co$ Golden Mean family of processes
		enables exploration of the consequences of finite Markov order $\mathscr{M}$ and finite cryptic order $\co \leq \mathscr{M}$.}
	\label{fig:MKGM}
\end{figure}

This 
$5$-$3$ Golden Mean process
is an example of the more general $\mathscr{M}$-$\co$ Golden Mean family of processes suggested by Fig.~\ref{fig:MKGM}.
For any such process, with the sole unknown transition parameter $p$,
we find that the stationary distribution over recurrent 
estimation states is 
$\stationary_T = \tfrac{1}{1+(\mathscr{M}+\co-1)p} \bigl[ 1, \, p, \, \dots \, , \, p \bigr]$
while the information vector over these recurrent states only has one nonzero entry
$\vec{f} = \bigl[ \tfrac{1}{p(1-p)}, \, 0, \, \dots \, , 0 \bigr]^\top$,
yielding an asymptotic information rate of 
\begin{align}
	\FIrate 
	&= 
	\Tpi \vec{f}
	= 
	\tfrac{1}{p (1-p) [1+(\mathscr{M}+\co-1)p] } ~.
\end{align}	

To determine the convergence towards this asymptotic information rate, 
we construct the estimation states and the transition probabilities among them via the MSP.
There are $\mathscr{M}+\co-1$ transient estimation states in total, with two branches coming out of the initial estimation state $\varepsilon(\varnothing)$: one branch of $\mathscr{M}-1$ ephemeral estimation states corresponding to the Markov order and another branch of $\co-1$ ephemeral estimation states corresponding to the cryptic order.

For this binary process,  the transient structure can be fully described by the following probabilities:
\begin{align}
\Pr(1) &= \tfrac{\mathscr{M} p}{1+(\mathscr{M}+\co-1)p } ~, \\
\Pr(0|1^n) &=  \tfrac{1}{\mathscr{M}-n+1 } & \text{for } 1 \leq n \leq \mathscr{M}-1  ~, \text{ and}\\ 
\Pr(1|0^n) &=  \tfrac{p}{1+(\co-n)p } & \text{for } 1 \leq n \leq \co -1  ~.
\end{align}
From these probabilites together with the recurrent structure of the process, the transition matrix $W$ and information vector $\ket{f}$ can be constructed.

Surprisingly, we observe that the cryptic order of this process, rather than its Markov order, determines when the myopic information rate reaches its asymptotic value.
For any $\mathscr{M}$-$\co$ Golden Mean process, 
the excess information saturates at the cryptic order
$\boldsymbol{\epsilon} = \EFI[\co]$, and
$\FIrate_{L} = \FIrate$ for all $L > \co$.

Based on this and other examples,
we conjecture more generally that, for any process of finite Markov order,
the information rate always saturates at the cryptic order (which is often less than the Markov order):
$\FIrate_{L} = \FIrate$ for all $L > \co$.
In the case of infinite Markov order,
the relationship between myopic information and cryptic order is more subtle.

\subsection{Infinite Markov order}

Typical 
processes with infinite Markov order never exactly reach the asymptotic information rate at any finite observation length.~\footnote{There are however exceptions for certain parametrizations of a process that yield a null or partially null information vector.}
In general, 
the net Fisher information provided by sequences of length $L$ is 
$\FI = L \FIrate +  \EFI[L] = \sum_{\ell = 1}^L \FIrate_\ell$.
Convergence properties of the
myopic information rate 
$\FIrate_\ell$
towards the asymptotic rate
$\FIrate$
depend directly on the modes of decay of the 
estimation-state metadynamic, which can be inferred from the mixed-state presentation.

In the simplest case of infinite Markov order, 
the myopic information rate converges 
as a sum of decaying exponentials.
We see an example of this in the Even process, 
shown in the fourth row of Table \ref{tab:ProcessTable}.

Despite a finite latent-state representation of only two hidden states,
the Even process 
has infinite Markov order.
This arises from the process' constraint that there can only be an even number of `1's between `0's:
Memory about the last zero, which could have occurred in arbitrarily distant history, 
must be retained 
to resolve the parity.
Nevertheless,
from the two-state unifilar presentation,
we immediately find the Fisher information rate
of the process if one aims to learn the transition parameter $p$.  
In this case, the asymptotic learning
rate is a scalar quantity:
\begin{align}
	\FIrate 
	&= 
\Tpi \vec{f}
	= 
	\tfrac{1}{p (1+p)(1-p) } ~.
\end{align}	

Convergence towards this asymptotic information rate (and the exact bound on estimator variance)
can be exactly quantified by constructing the estimation-state-to-state metadynamic.
Observations induce four estimation states for the Even process,
and the metadynamic among these is described by a diagonalizable but non-normal matrix $W$.
The modes of relaxation of this matrix $W$
exactly yield the convergence properties of the myopic information.
We encourage the reader to work through
App.~\ref{sec:Even}, 
which provides details of this derivation, and 
should serve as a useful tutorial.
In the end, we find that the myopic information rate 
oscillates around the asymptotic information rate 
as it decays exponentially towards it,
according to 
\begin{align}
	\FIrate_L 
	& = \begin{cases}
			\FIrate - \frac{p^{L/2}}{p(1+p)^2} & \text{for even } L \\
			\FIrate + \frac{p^{(L-1)/2}}{p(1+p)^2} & \text{for odd } L ~.
		\end{cases}      
\end{align}

Accumulating $\EFI = \sum_{\ell=1}^L ( \FIrate_\ell - \FIrate)$, 
we obtain the myopic excess information
\begin{align}
	\EFI = \frac{1}{p(1+p)^2} \bigl[ 1 - p^{L/2} \delta_{L/2 \in \mathbb{Z}} \bigr]
\end{align}
where $ \delta_{L/2 \in \mathbb{Z}}$ is equal to one if $L$ is an even number and is equal to zero otherwise.
Clearly,
$\boldsymbol{\epsilon} = \lim_{L \to \infty} \EFI =  \tfrac{1}{p(1+p)^2}$.

\subsection{Infinite Markov order; Finite symmetry-collapse index}
\label{sec:SymmetryCollapse}

The Teddy Bear process, named for the cute shape of its automaton, has a stationary distribution 
$\stationary_T = \tfrac{1}{1+2p+4q} \bigl[ 1, p, p, q, q, q, q \bigr]$
over its recurrent estimation states.
From the automaton, it is easy to see that there is parameter dependence only when leaving the `A' state; hence the information vector restricted to these recurrent states has a single non-zero entry for each pair of parameters $(p,p)$, $(p,q)$, or $(q,q)$.
The asymptotic information rate for this process is thus relatively 
easy to assemble from the labeled directed graph, 
with elements
\begin{align}
	\FIrate_{p,p}  &= \stationary_T  \vec{f}_{p,p} = \tfrac{1-q}{p(1-p-q)(1+2p+4q)} ~,\\
	\FIrate_{p,q} =  \FIrate_{q,p}  &= 
	\stationary_T  \vec{f}_{p,q} = \tfrac{1}{(1-p-q)(1+2p+4q)} 
	  ~, \text{ and}\\
	\FIrate_{q,q}  &= \stationary_T  \vec{f}_{q,q} = \tfrac{1-p}{q(1-p-q)(1+2p+4q)} ~,
\end{align}	
yielding
\begin{align}
	\FIrate = 
	\begin{bmatrix}
		\FIrate_{p,p} & \FIrate_{p,q} \\
		\FIrate_{q,p} & \FIrate_{q,q}  
	\end{bmatrix}
	= \gamma \begin{bmatrix}
		\tfrac{1-q}{p} & 1 \\
		1 & \tfrac{1-p}{q} 
	\end{bmatrix}
\end{align}	
where 
\begin{align}
	\gamma = \tfrac{1}{(1-p-q)(1+2p+4q)}~.
\end{align}

As always, the observation-induced convergence 
to this asymptotic information rate
is fully determined from the metadynamic among estimation states.
For this example process, we begin to see some interesting coexisting transient motifs.  In particular, if we inspect the directed graph structure for the estimation-state metadynamic, there is a leaky-period-three motif associated with resolving the phase of the 3-cycle; this is responsible for the process' infinite Markov order.  We see a coexisting ephemeral chain on the left-most part of the graph induced by sequential `2' observations, which is reminiscent of the structure that determined Markov order in the earlier $\mathscr{M}$-$\co$ Golden Mean process; here it determines the symmetry-collapse index $\nu_0(W)=3$, which is the size of $W$'s largest Jordan block associated with the eigenvalue of zero.
A shorter ephemeral chain due to sequential `0' observations
has length equal to the cryptic order of $\co=2$.

In the plot for the convergence of the myopic information,
we see ephemeral contributions from the zero eigenmode of the estimation-state metadynamic, up to length $L=2$, after which
the myopic information only deviates from its asymptotic value 
according to the decaying period-3 mode.
To demonstrate this more clearly, 
we isolated contributions from the  zero-eigenprojector $W_0$, which can be constructed relatively easily via
$W_0 = I - \sum_{\lambda \in \Lambda_W \setminus \{ 0 \}} W_\lambda $
since $W_\lambda = \ket{\lambda} \! \bra{\lambda} / \braket{ \lambda | \lambda}$ for $\lambda \neq 0$.
In particular,
we can always decompose
$\FIrate_L = \FIrate_L^{\multimap} + \FIrate_L^{\rightsquigarrow}  $,
where 
$(\FIrate_L^{\multimap} )_{m,n} \coloneqq \StartMS W_0 W^{L-1} \ket{f_{m,n}}$
includes only contributions from the zero-eigenmode
and 
$(\FIrate_L^{\rightsquigarrow} )_{m,n} \coloneqq \StartMS (I - W_0) W^{L-1} \ket{f_{m,n}} = \StartMS W^{\mathcal{D}} W^{L} \ket{f_{m,n}}$
contains the contributions from all other eigenmodes.
Here, $W^{\mathcal{D}} $ is the Drazin inverse of $W$~\cite{Riec18_Beyond}.
This parallels the decomposition of myopic entropy
introduced in Ref.~\cite{Riec18_SSAC2}.

In the myopic information plot for the Teddy Bear process shown in Table \ref{tab:ProcessTable}, we plot $(\FIrate_L^{\rightsquigarrow} )_{p,p}$ as black `$\times$'s, on top of the full myopic information rate $(\FIrate_L )_{p,p} = (\FIrate_L^{\multimap} )_{p,p} + (\FIrate_L^{\rightsquigarrow} )_{p,p}$ shown as magenta circles connected by the blue line.
We see that the zero eigenspace only contributes to the myopic information up to a sequence length of $L = \co = 2$
due to the process' depth-two cryptic ephemeral structure.

The symmetry-collapse index $\nu_0(W)$
always
sets the maximal sequence length 
at which the zero-eigenspace of $W$ contributes to 
the myopic information rate.
However, in this and other examples, we find that there is an even earlier cutoff of this contribution due to cryptic properties of the process.
This motivates us to
posit a more general form of our conjecture about cryptic order, which was earlier restricted to the case of finite Markov order.

For processes with finite cryptic order,
we conjecture that 
$\co$ 
sets the maximal sequence length 
at which the zero-eigenspace of $W$ contributes to 
the myopic information rate.

Much more generally still, we conjecture that the \emph{cryptic index}
of a process~\cite{Riec18_SSAC2}
sets the maximal sequence length 
at which the zero-eigenspace of $W$ contributes to 
the myopic information rate.
The cryptic index
is equal to the cryptic order whenever the latter is finite.

\subsection{No finite unifilar presentation}

The next element of complexity 
highlights the stark difference 
in resources required to
predict rather than simply generate a process.
The sixth row of Table \ref{tab:ProcessTable}
shows the Simple Nonunifilar Source (SNS):
A process with
a simple two-state generator,
but which has no finite unifilar presentation.
The labeled transition matrices
of the nonunifilar HMM,
$T^{(0)} 
= \begin{bmatrix} 1-p & p \\ 0 & 1-q \end{bmatrix}$
and 
$T^{(1)} 
= \begin{bmatrix} 0 & 0 \\ q & 0 \end{bmatrix}$,
sum to the net transition matrix 
$T =  \begin{bmatrix} 1-p & p \\ q & 1-q \end{bmatrix}$
from which we find the stationary distribution
$\stationary_T = \tfrac{1}{p+q} \bigl[ q, \, p \bigr]$.

From a spectral decomposition of the labeled transition matrices, for $p \neq q$,
we find 
\begin{align}
	\bigl( T^{(0)} \bigr)^n 
	&= \tfrac{(1-p)^n}{p-q}
	\begin{bmatrix} 1 \\ 0 \end{bmatrix}
	\begin{bmatrix} p \! - \! q, & -p \end{bmatrix}
	+
	\tfrac{(1-q)^n}{p-q}
	\begin{bmatrix} p \\ p \! - \! q \end{bmatrix}
	\begin{bmatrix} 0, & 1 \end{bmatrix}
\end{align}	
whereas 
\begin{align}
	\bigl( T^{(1)} \bigr)^n 
	&= I_2 \delta_{n,0} + q \delta_{n,1} 
	\begin{bmatrix} 0 \\ 1 \end{bmatrix}
	\begin{bmatrix} 1, & 0 \end{bmatrix}
\end{align}

This allows us to calculate transition probabilities among estimation states via 
construction
of the MSP.
The initial estimation state, before any observed history,
has transition probabilities 
$\Pro (0) = \tfrac{p+q - pq}{p+q}$
and 
$P_{\params}(1) = \tfrac{pq}{p+q}$
out of it.
Transition probabilities among all other estimation states
are given by
\begin{align}
\Pro \bigl( 0| \es(0^n) \bigr) 
	&=
	\tfrac{p^2 (1-q)^{n+1} - q^2 (1-p)^{n+1} }{p^2 (1-q)^n - q^2 (1-p)^n} ~, \\
\Pro \bigl(1| \es(0^n) \bigr) 
	&=
	\tfrac{p^2 q (1-q)^{n} - pq^2 (1-p)^{n} }{p^2 (1-q)^n - q^2 (1-p)^n } ~, \\
\Pro \bigl(0| \es(10^n) \bigr) 
	&=
	\tfrac{p (1-q)^{n+1} - q (1-p)^{n+1} }{ p (1-q)^{n} - q (1-p)^{n}  } ~, \text{ and} \\
\Pro \bigl(1| \es(10^n) \big) 
	&=
	\tfrac{p q (1-q)^{n} - p q (1-p)^{n} }{ p (1-q)^{n} - q (1-p)^{n}  } ~.
\end{align}	

Details are given in App.~\ref{sec:SNS}

As seen in the estimation-state metadynamic for the SNS in Table \ref{tab:ProcessTable},
observing a `1' always resets an observer to a particular 
estimation state.
However, chains of `0's, either before or after ever observing a `1'
lead to corresponding chains of estimation states---with one transient chain and one recurrent.

The resulting convergence to the asymptotic information rate is nearly (although not exactly) exponential
with increasing sequence length.

\subsection{No finite HMM} 
The Parentheses Matching process
shown in the next row of Table \ref{tab:ProcessTable}
has no finite HMM presentation, as proven in Ref.~\cite{Uppe97a}.
Nevertheless, the 
estimation states 
and the transitions among them
have a simple repetitive structure
that allows calculation of the myopic information up to arbitrary sequence length.
The nonstationarity of this process is not only manageable to analyze,
but even simplifies the analysis in this case.

\subsection{Non-ergodic} 
Nonergodic processes pose an interesting challenge for inference.
Each sample of length $L$ will come from some particular constituent process, but the constituent process associated with each sample will be chosen IID according to some distribution over models.
Learning the distribution over models and the constituent model parameters are interdependent when the constituent processes have overlapping support. 

In general the parametrization of a nonergodic process with $K$
disconnected components can be chosen such that 
\begin{align}
	\Pr_{\params}(X_{1:L}) = \sum_{k=1}^K g_k(\params_{\mathcal{M}}) \Pr_{\params_k}(X_{1:L}) ~,
\end{align}	
where the full set of parameters $\params$ 
naturally decompose into those $ \params_{\mathcal{M}} $ choosing the component and those $\params_k$ for each 
component. 

The transition matrices $T^{(x)}$ of a nonergodic process are block diagonal, with a block for each component.  This 
plays a prominent role in calculation of the MSP,
and leaves its signature in the 
Fisher information matrix.

Here, we briefly examine one of the simplest possible non-ergodic processes: that of two biased coins.
One coin, which outputs `1' with probability $p_1$
and outputs `0' otherwise,
will produce the sequence with probability $q$.
Another coin, which outputs `1' with probability $p_2$ and outputs `0' otherwise,
will produce the sequence with probability $1-q$.

In the belief-state metadynamic, $q$ enters the calculation as the prior $\boldsymbol{\mu} = [q, 1-q]$ about which coin to expect.
The labeled transition matrices 
$T^{(0)} = \begin{bmatrix}
	1-p_1 & 0 \\
	0 & 1-p_2
\end{bmatrix}$
and 
$T^{(1)} = \begin{bmatrix}
	p_1 & 0 \\
	0 & p_2
\end{bmatrix}$
are simultaneously diagonal and so commute.
The probability of any length-$L$ word with $n$ zeros
is thus the same, and induces the mixed state
$ \eta_{\params}^{(0^n 1^{L-n})} \propto 
 \boldsymbol{\mu} \bigl( T^{(0)} \bigr)^n  \bigl( T^{(1)} \bigr)^{L-n} 
=  \bigl[ q (1-p_1)^n p_1^{L-n}, \, (1-q)  (1-p_2)^n p_2^{L-n} \bigr] 
$,
such that 
\begin{align}
\Pro \bigl( 1 | \es (0^n 1^{L-n}) \bigr)
	&= \eta_{\params}^{(0^n 1^{L-n})} T^{(1)} \Tones \\
	&= \frac{q (1-p_1)^n p_1^{L-n+1} + (1-q)  (1-p_2)^n p_2^{L-n+1} }{ q (1-p_1)^n p_1^{L-n} + (1-q)  (1-p_2)^n p_2^{L-n} } ~.
\end{align}

As the length $L$ of a sequence increases,
the accessible layer of belief states marches 
successively down the infinite transient pyramid.
At each $L$ the probability distribution over mixed states
$\Pr_{\params}
\bigl( \ES_L = \es( 0^n 1^{L-n} ) \bigr)$ is the weighted sum of two binomial distributions.
Together, these elements lead to the surprising richness of the 
myopic information rate
observed in the last column of Table~\ref{tab:ProcessTable},
where the inset shows a very nontrivial observation-length dependence
through $L=250$.
The difficulty of learning from nonergodic processes 
would only grow more complicated with the nonergodic mixture of more complex components.

\section{Estimators, maximum likelihood, and independent samples}
\label{sec:MLE}

\begin{figure}
\includegraphics[width=0.465\columnwidth]{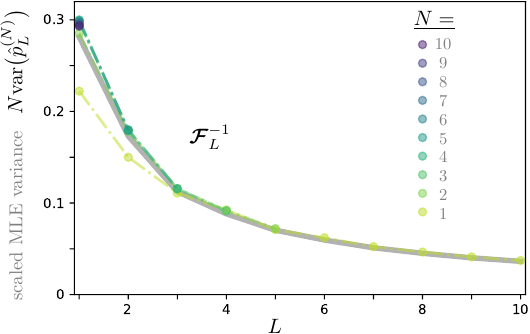}
	$\; \;$
	\includegraphics[width=0.465\columnwidth]{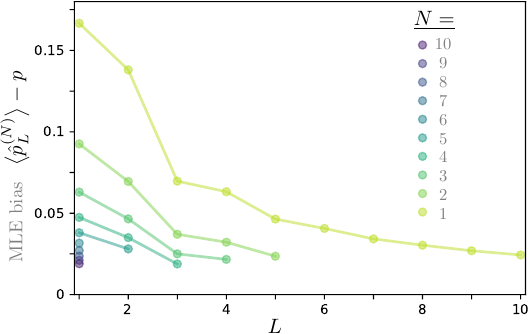}
	\caption{Left: 
		The Cramer--Rao lower bound on variance of any unbiased estimator of process parameters---$\FI^{-1}/N$---is achieved by the maximum-likelihood estimator (MLE) as either the number of samples $N$ or the length of each sample $L$ increases, for identifiable models of ergodic processes.  
		Here, we demonstrate this for the infinite-Markov-order Even process, at transition parameter $p=1/2$.
		Left:  Exact scaled variance $N \text{var}(\hat{\vec{\theta}}_L^{(N)})$, plotted together with $\FI^{-1}$ shown in thick gray.
Right:  Exact bias of the MLE, which reduces as either $N$ or $L$ increases.
	Even for a single sample ($N=1$) of an ergodic process, the MLE becomes unbiased and efficient as the observation length $L$ grows large.}
	\label{fig:Even_MLE}
\end{figure}	

The typical Cramer--Rao bound is stated in terms of unbiased estimators.
The maximum-likelihood estimator (MLE)---or, equivalently, the maximum-log-likelihood estimator---is a very commonly used estimator for model parameters.
However, the MLE is typically biased for a single finite-length sample of the process.
Nevertheless, 
it is well known that when 
many independent samples are pooled, the 
MLE is asymptotically unbiased and even asymptotically efficient---it achieves the Cramer--Rao lower bound as the number of independent samples 
grows large.

More precisely, for $N$ independent samples of the process, each of $L$ consecutive observations, the variance of the MLE $\hat{\vec{\theta}}_L^{(N)}$ attained from these net $LN$ observations converges in probability to a normal distribution:
\begin{align}
	\hat{\vec{\theta}}_L^{(N)}  & \to \mathcal{N}( \params, \, \FI^{-1}/N) & \text{as } N \to \infty
\end{align}	
where $\params$ is the true parameter of
an identifiable generative model,
and $\mathcal{N}(m,v)$ is a multivariate normal distribution with mean $m$ and covariance matrix $v$.
In particular,
$\lim_{N \to \infty}
N \text{var}(\hat{\params}_L^{(N)}) 
=
\FI^{-1} $.

Let's look again at the infinite-Markov-order Even process as an example
with unknown transition parameter $p$.
The
left half of
Fig.~\ref{fig:Even_MLE} shows how the variance of the MLE $\hat{p}_L^{(N)}$ of $N$ independent samples of length $L$ 
depends on both $N$ and $L$. 
This verifies that 
$\text{var} \bigl( \hat{p}_L^{(N)} \bigr)$ 
converges to the Cramer--Rao lower bound 
$ \FI^{-1}/N$
as either $N$ or $L$ grows large.
Notably,
we recognize that, even the MLE for a single sample (i.e., $N=1$) of an ergodic process becomes unbiased (see right half of Fig.~\ref{fig:Even_MLE}) and efficient 
(see left half of Fig.~\ref{fig:Even_MLE})
as the observation length increases, 
due to typicality.
Nonergodic processes in contrast require large $N$, even if $L$ is large.

In the real world,
there are distinct costs associated with
(i) extending a sequence 
(i.e., extending $L$)
or (ii) drawing a new independent sequence from the stochastic process
(i.e., increasing $N$).
These are two distinct ways of driving down the variance of parameter estimators, and thus present an interesting tradeoff.

As pointed out by Ref.~\cite{Radaelli22_Fisher},
it is not sufficient to simply ignore correlations in an attempt to obtain independent samples of short length---indeed, these samples are not independent.
Within a sequence, 
only distant parts of a correlated
process can be effectively independent due to decaying correlations.

It is notable that the Cramer--Rao lower bound for correlated stochastic processes is relatively simple to calculate using our methods at any $L$ and any $N$.
In contrast, the exact bias and variance of the maximum-likelihood estimator quickly becomes intractable for a large number of samples $N$ and/or large sequence length $L$ due to exponential growth of the support of the distribution, with $|\mathcal{X}|^{LN}$ elements.
Indeed, these overwhelming numerical demands 
practically limit $L$ for appreciable $N$ 
in Fig.~\ref{fig:Even_MLE}.
The fact that the variance of the biased MLE, $\text{var}\bigl( \hat{\params}_L^{(N)} \bigr)$, 
asymptotically and rather quickly converges to the Cramer--Rao lower bound 
$ \FI^{-1}/N$
can thus be seen as another rather powerful 
practical endorsement of our results.

\section{Overparametrization}
\label{sec:Overparam}

\begin{figure}
	\begin{minipage}{0.4\textwidth}
		\centering
		\vspace{-3em}
		\includegraphics[width=0.9\columnwidth]{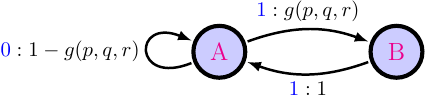}
\end{minipage}	
$\quad$
	\includegraphics[width=0.3\columnwidth]{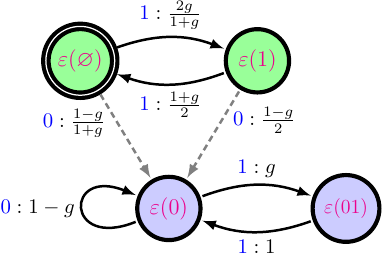}
	$\qquad \quad$
	\caption{A simple example of an overparametrized (i.e., non-identifiable) model, and the metadynamic among its estimation states.}
	\label{fig:Overparam}
\end{figure}	

Modern machine-learning models are typically highly degenerate, in that many parameter settings would lead to the same generative statistics.
This overparametrization has been associated with an ability to generalize, and is an important part of singular learning theory~\cite{Watanabe07_Almost}.
In such cases, the Fisher information matrix will have zero eigenvalues associated with the degenerate parameter settings, with the zero eigenspace representing a type of invariance or `symmetry'.

The methods developed here work just as well to infer the Fisher information matrix in these overparametrized cases, although the Cramer--Rao bound is no longer 
applicable.
The modes of the information matrix nevertheless still indicate modalities of inference that operate in parallel
with these degenerate subspaces.

As a basic example, we consider an overparametrized generator of the Even process, shown in the left half of Fig.~\ref{fig:Overparam}, where the single parameter $p$ has been replaced by a function 
$g(p,q,r)$ of three parameters
(e.g., the sum of three parameters $p+q+r$).
The parameters of this generative model are non-identifiable because of the freedom to choose any combination of the parameters that yield the same value of $g$ consistent with the generated stochastic process.
As shown in the right half of Fig.~\ref{fig:Overparam}, 
the estimation states are the same for this overparametrized model as they were for the minimal model examined earlier.
However, the resulting myopic information matrices, together with the resulting net Fisher information matrix $\FI$, are all singular due to the redundant parametrization.

The asymptotic matrix of Fisher information rates is
\begin{align}
	\FIrate = 
	\begin{bmatrix}
		\FIrate_{p,p} & \FIrate_{p,q} & \FIrate_{p,r} \\
		\FIrate_{q,p} & \FIrate_{q,q} & \FIrate_{q,r} \\
		\FIrate_{r,p} & \FIrate_{r,q} & \FIrate_{r,r} \\
	\end{bmatrix}
	= \gamma \begin{bmatrix}
		\partial_p g \\
		\partial_q g \\
		\partial_r g 
	\end{bmatrix}
	\begin{bmatrix}
		\partial_p g & \partial_q g & \partial_r g 
	\end{bmatrix} ~,
\end{align}	
where 
\begin{align}
	\gamma = \tfrac{1}{g(1 + g) (1-g)} ~.
\end{align}
This matrix of Fisher information rates
is thus a rank-1 symmetric matrix
with 
nonzero
eigenvalue
$\lambda = \gamma \vec{\nabla}_{\params} g \cdot \vec{\nabla}_{\params} g = 
\gamma \bigl[ (\partial_p g)^2 + (\partial_q g)^2 + (\partial_r g)^2  \bigr] $ 
associated with eigenvector $\begin{bmatrix}
		\partial_p g & \partial_q g & \partial_r g 
\end{bmatrix}^\top $,
and 
a twofold degenerate eigenvalue of 0
associated with the two-dimensional
degenerate
eigenspace
$a \begin{bmatrix}
  \partial_q g & - \partial_p g & 0 
\end{bmatrix}^\top 
+ 
b  \begin{bmatrix}
	0 &  \partial_r g & - \partial_q g
\end{bmatrix}^\top 
$
for any $(a,b) \in \mathbb{C}^2$.

Since the Fisher information matrix is singular, 
the simple Cramer--Rao bound is no longer 
applicable for 
constraining 
the variance of estimators.
However, this does not mean that the Fisher information matrix is useless.
Quite the opposite.
The 0-eigenspace reports the invariant subspace 
through which parameters can change without changing the process~\cite{Rothenberg71_Identification}.
The Drazin inverse of the 
Fisher information matrix, made up of the
complementary eigenspaces, plays a comparable role to 
the inverse of a non-degenerate Fisher information matrix---it bounds the variance of linear functionals of the parameters orthogonal to the invariant subspace.
Singular Fisher information matrices 
can also be employed 
via the 
alternative
Moore--Penrose pseudoinverse~\cite{Stoica01_Parameter}.

\section{Conclusion}
\label{sec:Conclusion}

How well can learning improve with increasing training data?
Here, we have answered this question in detail,
providing fundamental new results on the asymptotic rates 
of optimal learning and the rich transient structure 
of optimal learning with finite data.
In particular, our results 
show how the Fisher information of sequence probabilities 
scales with sequence length.
This lower bounds the variance of any unbiased estimator of 
the parameters of the process.
Moreover, the common maximum likelihood estimator (MLE) obtains this 
optimal learning rate asymptotically with increasing observation length.

Fisher information of correlated stochastic processes
has recently become a topic of interest 
in physical metrology---in both the classical and quantum regimes---for inference of physical parameters like temperature in the realistic scenario where a probe is measured sequentially~\cite{Radaelli22_Fisher, Smiga23_Stochastic, Boeyens23_Probe}.
In these initial explorations,
it has been noted that 
correlations may either enhance or hamper
metrological precision, depending on the particular system~\cite{Radaelli22_Fisher}. 
Our
closed-form expressions 
now
reveal 
how
either
sub- or super-additivity of the joint Fisher information
arises from eigen-contributions of the estimation-state metadynamic during learning.

Even for processes of infinite Markov order, our results provide
the information rate analytically from any unifilar presentation of the process.  
The exact bound on 
estimator
variance
can be calculated too, from the explicit metadynamic among estimation
states.

With this fundamental understanding of the ultimate limit of learning now in hand, 
it will be fascinating to apply and scale up 
the tools developed here to inspect the wildly successful recent AI developments like large language models (LLMs) that use transformer architectures~\cite{Vaswani17_Attention}.
It will be fruitful to compare the actual learning performance of these models with the optimal possible performance, as a function of 
increasing length of training data, for known processes 
that can serve as ground truth.
This calibration of neural networks'
learning performance 
would 
complement the benchmarking of
predictive performance from any particular learned model 
recently proposed
by Ref.~\cite{Marz23_Complexity}. 
Moreover,
our results
can inform the use of Baum--Welch inference,
Bayesian Structural Inference~\cite{Stre13a},
machine learning techniques---and indeed any inference technique---since 
our
exact 
bounds reveal how optimal inference depends on observed sequence length.

Our results can thus be used to benchmark
performance of any learning algorithm.
But these results teach us more.
More fundamentally,
we've learned---through estimation states
and their metadynamic---about
the structure of learning.

\section*{Acknowledgments}

I am grateful to Marco Radaelli, Alec Boyd, and Alexander Gietelink Oldenziel for stimulating discussion 
about the Fisher information of stochastic processes, which helped sharpen my thoughts on the topic.

\appendix

\section{Detailed derivation}
\label{sec:ddev}

Starting with Eq.~\eqref{eq:dF_explicit},
we turn the sum over histories 
into a sum over estimation states
using the simple fact that 
$\Pr_{\params}(X_L = x | X_{1:L-1} = w ) = \Pr_{\params} \bigl( X_L = x |  \ES_L = \es(w) \bigr)$. 
The myopic information rate then becomes
\begin{align}
	(\dF)_{m,n}
	&= - \!\! \sum_{w \in \mathcal{X}^{L-1}}
	\sum_{x \in \mathcal{X}}
	\Pr_{\params}(X_{1:L-1} = w )
	\Pr_{\params}(X_L = x | X_{1:L-1} = w ) \,
	\partial_{\theta_m} \partial_{\theta_n}   \ln \Pr_{\params}(X_L = x | X_{1:L-1} = w)  \\
	&= - \!\! \sum_{w \in \mathcal{X}^{L-1}}
	\sum_{x \in \mathcal{X}}
	\Pr_{\params}(X_{1:L-1} = w )
	\Pr_{\params}(X_L = x | \ES_L = \es(w) ) \,
	\partial_{\theta_m} \partial_{\theta_n}   \ln \Pr_{\params}(X_L = x |  \ES_L = \es(w) )  \\
	&= - \!\! \sum_{w \in \mathcal{X}^{L-1}}
	\sum_{x \in \mathcal{X}}
	\biggl[ \sum_{s \in \esSet}
	\Pr_{\params}(\ES_L = s, X_{1:L-1} = w ) \biggr]
	\Pr_{\params}(X_L = x | \ES_L = \es(w) ) \,
	\partial_{\theta_m} \partial_{\theta_n}   \ln \Pr_{\params}(X_L = x |  \ES_L = \es(w) )  \\
	&= - \!\!\!\! \sum_{w \in \mathcal{X}^{L-1}} \!
	\sum_{x \in \mathcal{X}}
\sum_{s \in \esSet}
	\underbrace{\Pr_{\params}(\ES_L \! = \! s | X_{1:L-1} \! = \! w ) }_{= \delta_{s,\es(w)}}
	\Pr_{\params}(X_{1:L-1} \! = \! w ) 
\Pr_{\params}(X_L \! = \! x | \ES_L \! = \! \es(w) ) \,
	\partial_{\theta_m} \! \partial_{\theta_n}   \ln \Pr_{\params}(X_L \! = \! x |  \ES_L \! = \! \es(w) )  \\
	&= - \!\! \sum_{w \in \mathcal{X}^{L-1}}
	\sum_{x \in \mathcal{X}}
\sum_{s \in \esSet}
	\delta_{s,\es(w)}
	\Pr_{\params}(X_{1:L-1} = w ) 
	\Pr_{\params}(X_L = x | \ES_L = s ) \,
	\partial_{\theta_m} \partial_{\theta_n}   \ln \Pr_{\params}(X_L = x |  \ES_L = s )  \\
	&= - 
	\sum_{x \in \mathcal{X}}
	\sum_{s \in \esSet}
	\biggl[
	\underbrace{ \sum_{w \in \mathcal{X}^{L-1}}
	\delta_{s,\es(w)}
	\Pr_{\params}(X_{1:L-1} = w ) }_{=\Pr_{\params} (\ES_L = s) } \biggr]
	\Pr_{\params}(X_L = x | \ES_L = s ) \,
	\partial_{\theta_m} \partial_{\theta_n}   \ln \Pr_{\params}(X_L = x |  \ES_L = s )  \\
&= - \!\! \sum_{s \in \esSet}
	\sum_{x \in \mathcal{X}}
	\Pr_{\params}(\ES_L = s )
	\Pr_{\params}(X_L = x | \ES_L = s) \,
	\partial_{\theta_m} \partial_{\theta_n}   \ln \Pr_{\params}(X_L = x | \ES_L = s)  \\
	&= - \!\! \sum_{s \in \esSet}
	 \StartMS W^{L-1} \ket{s}
	\sum_{x \in \mathcal{X}}
	\Pr_{\params}(X_L = x | \ES_L = s) \,
	\partial_{\theta_m} \partial_{\theta_n}   \ln \Pr_{\params}(X_L = x | \ES_L = s)  \\
&= \StartMS W^{L-1} \ket{ f_{m,n} } 
	~,
\end{align}	
for any integer $L \geq 1$, where
we have introduced the $L$-independent information vector
\begin{align}
	\ket{ f_{m,n} } = 
	- \!\! \sum_{s \in \esSet }
	\sum_{x \in \mathcal{X}}
	\ket{s}
	\Pro( x | s) \,
	\partial_{\theta_m} \partial_{\theta_n}   \ln \Pro( x |  s)  
~.
\end{align}	
In Sec.~\ref{sec:HMMs}, we point out that the
conditional probability 
$\Pro( x |  s)  $
can be calculated via mixed states as 
$\Pro( x |  s)  
=
\eta_{\params}^{(w)} T^{(x)} \Tones $ for any $w \in s$.

\section{A pedagogical example: the Even Process}
\label{sec:Even}

\begin{figure}[b]
	\includegraphics[width=0.25\columnwidth]{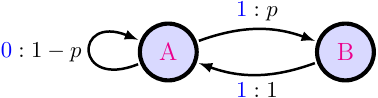}
	$\qquad$
	\includegraphics[width=0.25\columnwidth]{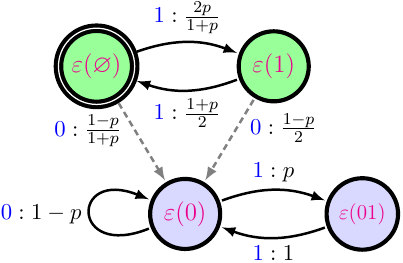}
	\caption{Left: A unifilar presentation of the Even process, parametrized by the unknown transition probability $p \in (0,1)$, from which the Fisher information rate can be directly and immediately calculated.  This process has infinite Markov order.  Right: The estimation-state metadynamic
of the Even process, from which the myopic information rate and excess information can be calculated.  These quantities bound the variance of any estimator of the stochastic process.}
	\label{fig:Even}
\end{figure}	

As a demonstration of our methods and results,
we thoroughly analyze the Even Process shown in Fig.~\ref{fig:Even} which,
despite a finite latent-state representation,
has infinite Markov order.
From the unifilar presentation in Fig.~\ref{fig:Even}(a),
we immediately find the Fisher information rate
of the process, if one aims to infer the transition parameter $p$.  
From the latent-state-to-state transition matrix
$T = \begin{bmatrix}
	1-p & p \\
	1 & 0
\end{bmatrix}$,
we find 
$\Tpi = \Tpi T
= \tfrac{1}{1+p} 
\begin{bmatrix}
	1 & p 
\end{bmatrix}$.
The information vector, restricted to these recurrent states, is 
$
\vec{f}
= 
\begin{bmatrix}
	\frac{1}{p(1-p)} & 0
\end{bmatrix}^\top$.
In this case, the information rate is a scalar quantity:
\begin{align}
	\FIrate 
	&= 
\Tpi \vec{f}
	= \bigl[ p (1+p)(1-p) \bigr]^{-1}~.
\end{align}	
This gives the asymptotic rate of the growth of Fisher information for this process as longer sequences are considered.
Notably the information rate diverges
as the transition rate approaches either extreme of 0 or 1.
In the middleground, where Fisher information is low,
estimates of the transition parameter must vary most.

To exactly bound the variance of an estimator
requires the 
metadynamic among estimation states, which we calculate via the
MSP of the process.
To obtain the mixed-state presentation,
we utilize the labeled transition matrices
$T^{(0)} = \begin{bmatrix}
	1-p & 0 \\
	0 & 0
\end{bmatrix}$
and 
$T^{(1)} = \begin{bmatrix}
	0 & p \\
	1 & 0
\end{bmatrix}$.
The exact bound on variance will depend on the initial distribution over model states, which could be non-stationary in general.  
Here we will analyze the stationary process, which implies that the process starts in the stationary distribution $\Tpi$.
We then calculate the four unique mixed states
$\eta_{\params}^{(\varnothing)} = \Tpi$,
$\eta_{\params}^{(1)} = [\tfrac{1}{2}, \tfrac{1}{2}]$,
$\eta_{\params}^{(0)} = [1, 0]$,
and 
$\eta_{\params}^{(01)} = [0, 1]$,
which are in one-to-one correspondence with the estimation states.
Using these mixed states, we calculate the
net transition matrix among 
estimation states
to be
\begin{align}
	W = 
	\begin{bmatrix}
		0 & 2p/(1+p) & (1-p)/(1+p) & 0 \\
		(1+p)/2 & 0 & (1-p)/2 &  0 \\
		0 & 0 & 1-p & p \\
		0 & 0 & 1 & 0
	\end{bmatrix} ~.
\end{align}
On this larger space of 
estimation states, we use $W$ with Eq.~\eqref{eq:BinarySimplification2} to calculate that
the information vector is 
\begin{align}
	\ket{f} = \bigl[ \tfrac{2}{  p (1-p) (1+p)^2} , \, \tfrac{1}{(1+p)(1-p)} , \, \tfrac{1}{p(1-p)} , \, 0  \bigr]^\top ~.
\end{align}

Recall that the myopic information rate 
is given by 
$\FIrate_L = \braket{\es(\varnothing) | W^{L-1} | f }$,
where $\bra{\es(\varnothing)} = [1 , 0 , 0, 0 ]$.
In principle, we now have all the parts, but a spectral decomposition of $W$ lets us go a step further to obtain an exact analytic expression for 
the myopic information rate.

The transition matrix $W$ is a diagonalizable but non-normal matrix with eigenvalues $\Lambda_W = \bigl\{ 1 , \, -p, \, \pm \sqrt{p} \bigr\}$,
and corresponding eigenmatrices
\begin{align}
	W_\lambda = \ket{\lambda} \! \bra{\lambda} / \braket{\lambda| \lambda} ~,
\end{align}
with left eigenvectors 
\begin{align}
	\bra{1} &= \Wpi = \bigl[ 0, \, 0, \, \tfrac{1}{1+p}, \, \tfrac{p}{1+p} \bigr] ~, \\
	\bra{-p} &= [0, \, 0 , \, -1 , \, 1 ] ~, \text{ and} \\
	\bra{\pm \sqrt{p}} &= \bigl[  \tfrac{1+p}{p \pm \sqrt{p}} , \, \tfrac{2}{1 \pm \sqrt{p}} , \, \tfrac{\mp \sqrt{p}}{p} , \, 1 \bigr] ~,
\end{align}
and right eigenvectors 
\begin{align}
	\ket{1} &= \Wones = [1, \, 1, \, 1, \, 1]^\top ~, \\
	\ket{-p} &=
	\bigl[ 0 , \, \tfrac{1-p}{2} , \, -p , \, 1 \bigr]^\top  ~, \text{ and} \\  
	\ket{\pm \sqrt{p}} &= \bigl[ \pm \sqrt{p}, \, \tfrac{ 1+p }{ 2}, \, 0, \, 0 \bigr]^\top
	~.
\end{align}

We can now calculate the myopic information rate as 
\begin{align}
	\FIrate_L 
	&= 
	\braket{\es(\varnothing) | W^{L-1} | f } \\
	&= \sum_{\lambda \in \Lambda_W} \lambda^{L-1} 
	\braket{\es(\varnothing) | \lambda \rangle \langle \lambda | f } / \braket{\lambda| \lambda} \\
	&= \braket{\stationary| f} 	
	+ \sum_{\lambda \in \{ \pm \sqrt{p} \} } \lambda^{L-1} 
	\braket{\es(\varnothing) | \lambda \rangle \langle \lambda | f } / \braket{\lambda| \lambda} \\
	&= \FIrate + \frac{1-p}{2p(1+p)^2}
	\sum_{\lambda \in \{ \pm \sqrt{p} \} } \frac{\lambda^L}{\lambda+p}  
	\label{eq:Even_fL_WithSimpleEigSum}
	\\
	& = \begin{cases}
		\FIrate - \frac{p^{L/2}}{p(1+p)^2} & \text{for even } L \\
		\FIrate + \frac{p^{(L-1)/2}}{p(1+p)^2} & \text{for odd } L ~.
	\end{cases}      
\end{align}
This tells us that, 
for the infinite-Markov-order Even process,
the myopic information rate 
oscillates around the asymptotic information rate 
as it decays exponentially towards it.

To observe the implications for the net Fisher information in length-$L$ sequences, recall that 
$\FI = \sum_{\ell = 1}^{L} \FIrate_\ell$.
Thus, from Eq.~\eqref{eq:Even_fL_WithSimpleEigSum},
we find
\begin{align}
	\FI 
	&=  \sum_{\ell = 1}^{L} \FIrate_\ell \\
	&= L \FIrate + \frac{1-p}{2p(1+p)^2}
	\sum_{\lambda \in \{ \pm \sqrt{p} \} } \frac{ \sum_{\ell = 1}^L \lambda^\ell}{\lambda+p}  \\
	&=  L \FIrate + \frac{1}{2p(1+p)^2}
	\sum_{\lambda \in \{ \pm \sqrt{p} \} } ( 1-\lambda^L ) \\
	& =  L \FIrate + \frac{1 - p^{L/2} \delta_{L/2 \in \mathbb{Z}}}{p(1+p)^2} 
	\label{eq:Even_FI_LFplusEFI}
	\\
	&= \tfrac{1}{p (1+p) (1-p)} \bigl[ L + \tfrac{1-p}{1+p} ( 1 - p^{L/2} \delta_{L/2 \in \mathbb{Z}} ) \bigr] ~,
\end{align}
where $ \delta_{L/2 \in \mathbb{Z}}$ is equal to one if $L$ is an even number and is equal to zero otherwise.

From Eq.~\eqref{eq:Even_FI_LFplusEFI},
we can identify the myopic excess information
\begin{align}
\EFI = \frac{1}{p(1+p)^2} \bigl[ 1 - p^{L/2} \delta_{L/2 \in \mathbb{Z}} \bigr]
\end{align}	
and the asymptotic excess information
\begin{align}
	\boldsymbol{\epsilon} = \frac{ 1 }{p(1+p)^2} ~.
\end{align}	

The variance of any unbiased estimator $\hat{p}_L$ of the transition parameter $p$ of the Even process, based on length-$L$ observation sequences, is thus lower bounded by
\begin{align}
	\text{var}(\hat{p}_L) \geq \FI^{-1} 
	= p (1+p ) (1-p)
	\bigl[ L + \tfrac{1-p}{1+p} ( 1 - p^{L/2} \delta_{L/2 \in \mathbb{Z}} ) \bigr]^{-1}
~.
\end{align}	

Recall that the maximum likelihood estimator (MLE) is biased.  Nevertheless the MLE, $\hat{p}_L^{(N)}$, is asymptotically unbiased and efficient, in that it asymptotically achieves the Cramer--Rao lower bound with increasing number $N$ of independent samples of sequence-length $L$:
\begin{align}
	\lim_{N \to \infty}
	N \text{var}(\hat{p}_L^{(N)}) 
=
	 \FI^{-1} 
	= 
p (1+p ) (1-p)
	\bigl[ L + \tfrac{1-p}{1+p} ( 1 - p^{L/2} \delta_{L/2 \in \mathbb{Z}} ) \bigr]^{-1}
~.
\end{align}

\section{Simple nonunifilar source}
\label{sec:SNS}

The labeled transition matrices
of the two-state nonunifilar HMM for the SNS,
$T^{(0)} 
= \begin{bmatrix} 1-p & p \\ 0 & 1-q \end{bmatrix}$
and 
$T^{(1)} 
= \begin{bmatrix} 0 & 0 \\ q & 0 \end{bmatrix}$,
sum to the net transition matrix 
$T =  \begin{bmatrix} 1-p & p \\ q & 1-q \end{bmatrix}$
from which we find the stationary distribution
$\stationary_T = \tfrac{1}{p+q} \bigl[ q, \, p \bigr]$.

From a spectral decomposition of the labeled transition matrices, for $p \neq q$,
we find 
\begin{align}
	\bigl( T^{(0)} \bigr)^n 
	&= \tfrac{(1-p)^n}{p-q}
	\begin{bmatrix} 1 \\ 0 \end{bmatrix}
	\begin{bmatrix} p \! - \! q, & -p \end{bmatrix}
	+
	\tfrac{(1-q)^n}{p-q}
	\begin{bmatrix} p \\ p \! - \! q \end{bmatrix}
	\begin{bmatrix} 0, & 1 \end{bmatrix}
\end{align}	
whereas 
\begin{align}
	\bigl( T^{(1)} \bigr)^n 
	&= I_2 \delta_{n,0} + q \delta_{n,1} 
	\begin{bmatrix} 0 \\ 1 \end{bmatrix}
	\begin{bmatrix} 1, & 0 \end{bmatrix} 
	~.
\end{align}

This allows us to calculate transition probabilities among estimation states via construction
of the MSP.
The initial estimation state, before any observed history,
has transition probabilities 
$P_{\params}(0) = \tfrac{p+q - pq}{p+q}$
and 
$P_{\params}(1) = \tfrac{pq}{p+q}$
out of it.
With slight abuse of notation,
we will denote the information vector as 
$\ket{f_{p,p}}$,
$\ket{f_{q,q}}$,
and 
$\ket{f_{p,q}}$.
Then 
$\braket{\varepsilon_{\params}(\varnothing) | f_{p,p} } = \tfrac{q^3}{p (p+q)^2 (p+q-pq)} $.

A countably infinite string of transient estimation states are induced by only-zero histories.
We find
\begin{align}
	\stationary_T \bigl( T^{(0)} \bigr)^n 
	&= \tfrac{q (1-p)^n}{(p+q)(p-q)}
	\begin{bmatrix} p \! - \! q, & -p \end{bmatrix}
	+
	\tfrac{p^2 (1-q)^n}{(p+q)(p-q)}
	\begin{bmatrix} 0, & 1 \end{bmatrix}
\end{align}	
and 
\begin{align}
	\stationary_T \bigl( T^{(0)} \bigr)^n \Tones
	&= 
	\tfrac{p^2 (1-q)^n - q^2 (1-p)^n}{(p+q)(p-q)}
	~,
\end{align}	
from which we obtain the transition probabilities
\begin{align}
	\Pr(0|0^n) 
	&= \tfrac{\stationary_T \bigl( T^{(0)} \bigr)^{n+1} \Tones}{\stationary_T \bigl( T^{(0)} \bigr)^n \Tones} \\
	&=
	\tfrac{p^2 (1-q)^{n+1} - q^2 (1-p)^{n+1} }{p^2 (1-q)^n - q^2 (1-p)^n} & \text{for } n \geq 1 ~.
\end{align}	

Observing a `1' always resets the observer to the recurrent belief state
$\eta_{\params}^{(1)} = [1, \, 0]$.
Another countably infinite set of estimation states---this time recurrent---ensues from histories of reset followed by zeros.
We find
\begin{align}
	[1, \, 0] \bigl( T^{(0)} \bigr)^n 
	&= \tfrac{ (1-p)^n}{p-q}
	\begin{bmatrix} p \! - \! q, & -p \end{bmatrix}
	+
	\tfrac{p (1-q)^n}{p-q}
	\begin{bmatrix} 0, & 1 \end{bmatrix}
\end{align}	
and 
\begin{align}
	[1, \, 0]  \bigl( T^{(0)} \bigr)^n \Tones
	&= 
	\tfrac{p (1-q)^n - q (1-p)^n}{p-q}
	~,
\end{align}	
from which we obtain the transition probabilities
\begin{align}
	\Pr(0| 1 0^n ) 
	&= 
	\tfrac{ [1, \, 0]  \bigl( T^{(0)} \bigr)^{n+1} \Tones}{ [1, \, 0]  \bigl( T^{(0)} \bigr)^n \Tones} \\
	&=
	\tfrac{p (1-q)^{n+1} - q (1-p)^{n+1} }{p (1-q)^n - q (1-p)^n} & \text{for } n \geq 1 ~.
\end{align}	
These are all the pieces we need to calculate the full
information vector.

To obtain the asymptotic information rate, 
observe that the stationary probability of the 
recurrent
estimation state $\braket{\stationary | \es(1) } = \Pr_{\params}(1) = \tfrac{pq}{p+q}$,
while each subsequent recurrent state inherits some of this probability according to 
$\braket{\stationary | \es(1 0^n) } = \braket{\stationary | \es(1 0^{n-1}) } \Pr(0 | 10^{n-1}) $.
Hence
\begin{align}
	\braket{\stationary | \es(1 0^n) } 
	&= 
	\tfrac{pq}{p+q}
	\prod_{m=1}^{n-1}
	\Pr(0 | 10^{m}) \\
	&= 
	\tfrac{pq}{p+q}
	\prod_{m=1}^{n-1}
	\tfrac{p (1-q)^{n+1} - q (1-p)^{n+1} }{ p (1-q)^{n} - q (1-p)^{n}  } \\
	&= 
	\tfrac{pq \bigl[ p(1-q)^n - q(1-p)^n \bigr] }{(p+q)(p-q)}
	~.
\end{align}

From $\Pr(0|10^n) = u_n/g_n$ alone,
we can calculate the information vector restricted 
to the recurrent states, via Eq.~\eqref{eq:QuotientRule_Simplification1}.
The asymptotic information rate $\FIrate_{p,p}$ 
is then
\begin{align}
	\FIrate_{p,p} 
	&= \braket{\stationary | f_{p,p} } \\
	&=
	\sum_{n=0}^\infty
	\braket{\stationary | \es(1 0^n) } 
	\braket{\es(1 0^n) | f_{p,p} } \\
	&= 
	\sum_{n=1}^\infty
	\braket{\stationary | \es(1 0^n) } 
	\frac{\Bigl[ \bigl( \partial_{p} u_n \bigr) g_n -  u_n \bigl( \partial_{p} g_n \bigr)  \Bigr]^2 }{ g_n^2 u_n (g_n-u_n) } \\
	&= 
	\tfrac{pq }{(p+q)(p-q)}
	\sum_{n=1}^\infty
	\frac{\Bigl[ \bigl( \partial_{p} u_n \bigr) g_n -  u_n \bigl( \partial_{p} g_n \bigr)  \Bigr]^2 }{ g_n u_n (g_n-u_n) } ~,
	\label{eq:SNSFIrate_infinitesum}
\end{align}
where
\begin{align}
	u_n &= p (1-q)^{n+1} - q (1-p)^{n+1} ~, \\
	g_n &= p (1-q)^n - q (1-p)^n ~, \\
	\partial_{p} u_n &= (1-q)^{n+1} + (n+1) q (1-p)^n ~,  \\
	\partial_{p} g_n &= (1-q)^n + n q (1-p)^{n-1} ~, 
	\text{ and} \\
	g_n - u_n &= pq \bigl[ (1-q)^n - (1-p)^n \bigr] ~.
\end{align}

Eq.~\eqref{eq:SNSFIrate_infinitesum}
can be approximated 
arbitrarily well by truncating the sum at sufficiently large $n$.
Indeed, 
$\braket{\es(1 0^n) | f_{p,p} }$
approaches zero with large $n$
while 
$\braket{\stationary | \es(1 0^n) } $
converges to zero 
exponentially as $n$ increases.

\section{Relation to Blackwell's expression for entropy rate}
\label{sec:BlackwellAnalog}

If one starts with a nonunifilar presentation with latent states $\SSet$, 
even if there is no finite unifilar presentation,
the  Fisher-information 
rate can be expressed more generally as a Lebesgue integral over 
the simplex $\boldsymbol{\Delta}_{\SSet}$ of model-state probabilities, 
\begin{align}
	(\FIrate)_{m,n} 
	&= 
	- \int_{\eta \in \boldsymbol{\Delta}_{\SSet}}
	d \Blackwell(\eta)
	\sum_{x \in \Abet}
	\Pro(x| \eta ) \,
	\bigl[
	\partial_{\theta_m} \partial_{\theta_n}   \ln \Pr_{\params}(x | \overleftarrow{x}  ) 
	\bigr]_{\overleftarrow{x} \in \{ w \in \mathcal{X}^*  :\eta_{\params}^{(w)} = \eta \}}~,
	\label{eq:LikeBlackwell}
\end{align}	
weighing contributions via
the stationary distribution (a.k.a.\ the `Blackwell measure') $\Blackwell$
over observation-induced belief states $\eta$.
With slight abuse of notation,
we have used the shorthand $\Pro(x | \eta ) = \Pro(X_t = x | \St_t \sim \eta ) = \sum_{s \in \SSet} \eta(s) \Pro(X_t = x | \St_t = s )  $.
Eq.~\eqref{eq:LikeBlackwell} is analogous to Blackwell's expression for the entropy rate of a stochastic process~\cite{Blac57b}.

However, 
since $\eta_{\params}$ represents the model-state distribution induced by an equivalence class of histories,
the partial derivative
$\partial_{\theta_n} \eta_{\params}$ 
is typically nonzero. 
This complicates
the use of Eq.~\eqref{eq:LikeBlackwell};
Indeed, the following tempting 
expression
fails to supply the information rate:
\begin{align}
	(\FIrate)_{m,n} 
	&\neq 
	- \int_{\eta \in \boldsymbol{\Delta}_{\SSet}}
	d \Blackwell(\eta)
	\sum_{x \in \Abet}
	\Pro (x| \eta ) \,
\partial_{\theta_m} \partial_{\theta_n}   \ln \Pro (x | \eta  ) 
\end{align}	
since 
$\partial_{\theta_n} \eta$ must be zero for a fixed dummy vector $\eta$.

\end{document}